\documentclass[10pt,twocolumn,letterpaper]{article}

\usepackage[pagenumbers]{wacv} % To force page numbers, e.g. for an arXiv version

\usepackage{graphicx}
\usepackage{amsmath}
\usepackage{amssymb}
\usepackage{booktabs}

\usepackage[textsize=tiny]{todonotes}
\usepackage{arydshln}
\usepackage{amsmath}
\usepackage{amssymb}
\usepackage{mathtools}
\usepackage{amsthm}
\usepackage{amsmath}
\usepackage{dsfont}
\usepackage{multirow}
\usepackage{cite}
\DeclareMathOperator*{\argmax}{arg\,max}

\usepackage{algorithm}
\usepackage{algorithmic}

\usepackage{xcolor}
\newcommand{\colorname}{black}
\usepackage[pagebackref,breaklinks,colorlinks]{hyperref}

\usepackage[capitalize]{cleveref}
\crefname{section}{Sec.}{Secs.}
\Crefname{section}{Section}{Sections}
\Crefname{table}{Table}{Tables}
\crefname{table}{Tab.}{Tabs.}

\def\wacvPaperID{1006} % *** Enter the WACV Paper ID here
\def\confName{WACV}
\def\confYear{2025}

\newtheorem{theorem}{Theorem}[section]

\newtheorem{lemma}[theorem]{Lemma}

\begin{document}

%%%%%%%%% TITLE - PLEASE UPDATE
\title{CLIPScope: Enhancing Zero-Shot OOD Detection with Bayesian Scoring}

\author{Hao Fu, Naman Patel, Prashanth Krishnamurthy, Farshad Khorrami \\
{\tt\small \{hf881, nkp269, prashanth.krishnamurthy, khorrami\}@nyu.edu} \\
Department of Electrical and Computer Engineering\thanks{This paper is supported in part by the Army Research Office under grant number W911NF-21-1-0155 and by the New York University Abu Dhabi (NYUAD) Center for Artificial Intelligence and Robotics, funded by Tamkeen under the NYUAD Research Institute Award CG010.} \\
New York University, Brooklyn, NY, 11201 \\
}
\maketitle

%%%%%%%%% ABSTRACT
\begin{abstract}
Detection of out-of-distribution (OOD) samples is crucial for safe real-world deployment of machine learning models. Recent advances in vision language foundation models have made them capable of detecting OOD samples without requiring in-distribution (ID) images. However, these zero-shot methods often underperform as they do not adequately consider ID class likelihoods in their detection confidence scoring. Hence, we introduce CLIPScope, a zero-shot OOD detection approach that normalizes the confidence score of a sample by class likelihoods, akin to a Bayesian posterior update. Furthermore, CLIPScope incorporates a novel strategy to mine OOD classes from a large lexical database. It selects class labels that are farthest and nearest to ID classes in terms of CLIP embedding distance to maximize coverage of OOD samples. We conduct extensive ablation studies and empirical evaluations, demonstrating state of the art performance of CLIPScope across various OOD detection benchmarks. Code is available at \url{https://github.com/fu1001hao/CLIPScope}.
\end{abstract}

\section{Introduction}
\label{sec:introduction}

Machine learning systems often encounter challenges when dealing with out-of-distribution (OOD) samples, which are not represented in the training distribution. These systems are built on the assumption that the data encountered during testing will mirror the training distribution. OOD detection~\cite{YZLL21} is vital for reliable deployment of machine learning systems. This detection process employs a confidence scoring mechanism, in alignment with prior work, that assesses whether a data point is part of the known training distribution or an OOD instance. It designates higher values to in-distribution (ID) samples and lower values to OOD samples, thus enabling their detection. 

Traditional OOD detection methods predominantly focus on image data, overlooking the potential benefits of incorporating textual information. The concept of zero-shot OOD detection, which utilizes both textual and image information, was introduced by ZOC \cite{ELRS22}. This approach involves using image captioning to generate potential OOD labels for input instances, followed by a zero-shot classification using CLIP \cite{RKH21} with both ID and generated OOD labels. A noted limitation of ZOC, as highlighted by \cite{WLYL23} and \cite{NegLabel}, is its reduced effectiveness in generating relevant OOD labels for large ID class datasets, such as ImageNet-1K, resulting in decreased OOD detection accuracy. MCM \cite{MCGSLL22} addresses OOD detection by employing the maximum logit of scaled softmax as a confidence score. However, MCM's reliance solely on ID class labels, without fully leveraging open-world textual information, limits its performance, particularly with challenging OOD samples, as reported by \cite{WLYL23}. CLIPN \cite{WLYL23} introduces an advancement by incorporating a CLIP-based negative-text encoder trained with additional datasets, including 3 million image-text pairs. This encoder is designed to understand negative prompts. For an input categorized into a specific ID class, CLIPN calculates the confidence scores using both the negative-text encoder and the original CLIP text encoder, with the final OOD confidence score being the product of these two.  NegLabel \cite{NegLabel} proposes a mining algorithm for extracting candidate OOD labels and a grouping strategy to compute the OOD score. However, the performance of NegLabel is sensitive to the size of negative label space and the percentile distance used in the mining procedure. This work proposes a method that is robust against these parameters.

\begin{figure*}[ht]
    \centering
    \includegraphics[width=0.9\textwidth]{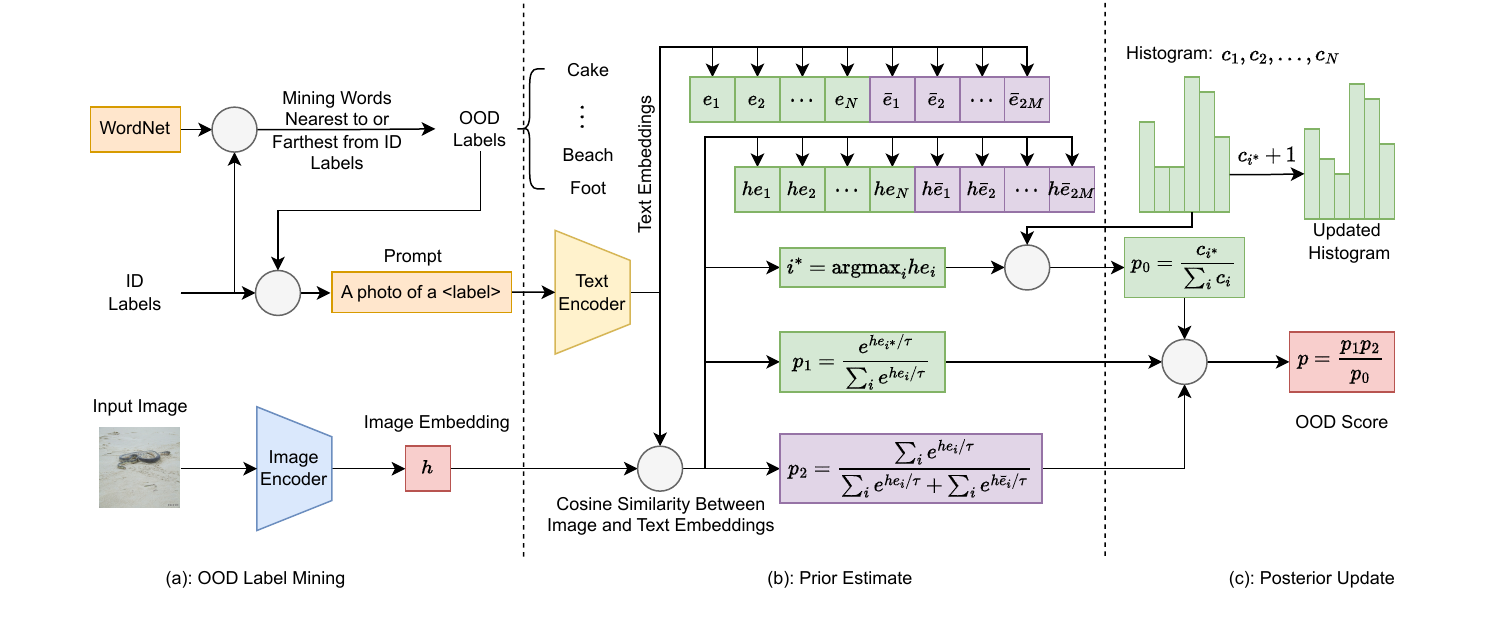}
    \caption{ The process of CLIPScope  involves three key stages: OOD label mining, prior estimate, and posterior update.
    }
    \label{fig:pipeline}
    \vspace{-0.6cm}
\end{figure*}

In this paper, we introduce CLIPScope, a novel zero-shot method that leverages Bayesian inference to enhance confidence scoring for OOD detection. The model evidence or marginal likelihood for our Bayesian scoring is calculated based on the class membership likelihood, determined by the proportion of number of instances in specific ID classes to prior instances. CLIPScope operates on the principle that without prior knowledge of OOD sample distribution, the chance of an instance being OOD should be consistent across ID classes. Therefore, classes with a higher frequency of instances are deemed more likely to include OOD samples. CLIPScope adjusts confidence scores accordingly, reducing them for OOD instances in these "high-frequency" classes while keeping scores for ID instances high. Furthermore, CLIPScope introduces a novel approach for mining potential OOD labels from WordNet lexical database~\cite{F98} by considering both the closest and farthest words to ID labels. This strategy aims to maximize the coverage of the OOD sample space, thereby enhancing the robustness and effectiveness of zero-shot OOD detection, in contrast to methods like NegMining~\cite{NegLabel} that only consider the most distant words from ID labels. After all, OOD samples close to ID space are more challenging to detect than OOD samples far from ID space. To address the issue that certain nearest OOD labels may fall within ID label space by being synonyms of ID labels, CLIPScope uses another two components that do not require any mined OOD labels.

\begin{figure}
    \centering
    \includegraphics[width=0.8\linewidth]{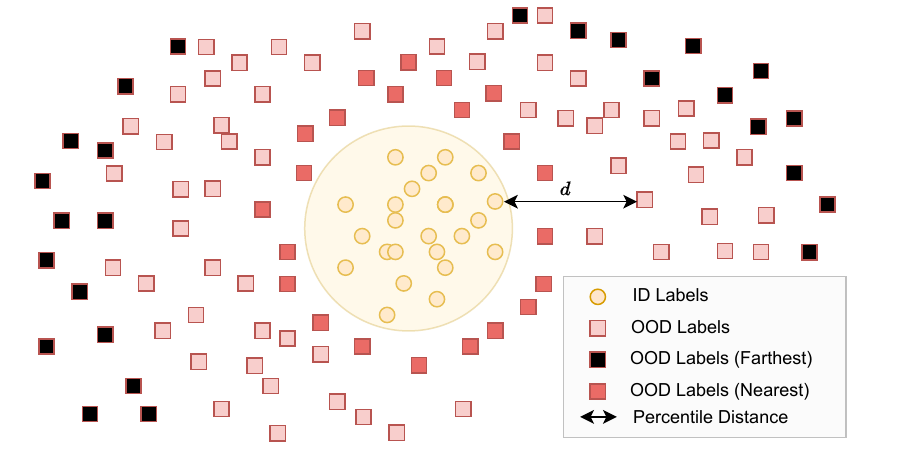}
    \caption{The nearest (red) and farthest (dark) labels are picked.}
    \vspace{-0.7cm}
    \label{fig:mining}
\end{figure}

The CLIPScope framework for OOD detection, illustrated in Fig.~\ref{fig:pipeline} begins by mining potential OOD labels from a vocabulary database like WordNet \cite{F98}, based on their lexical proximity to ID labels. This proximity is determined by the text embeddings of the words, with CLIPScope calculating the negative cosine similarities between each word's embedding and the entire ID label space. Next, CLIPScope identifies the 5th percentile of these similar words for each ID label and selects the top M farthest words and top M nearest words as potential OOD labels, as illustrated in Fig.~\ref{fig:mining}. This mining procedure conducted prior to the inference phase, establishes the OOD labels, which are thereafter fixed and do not require updates in subsequent operations. 
Next, using identified ID and OOD label text embeddings, and extracted image embedding for a given input, CLIPScope estimates the prior likelihood for an input being ID. It first identifies the ID class, $i^*$, closest to the image embedding, and then calculates $p_1$, the confidence that the input belongs to class $i^*$, and $p_2$, the confidence that a given input belongs to any ID class, as shown in Fig.~\ref{fig:pipeline}(b). The final step given in Fig.~\ref{fig:pipeline}(c) involves updating the posterior with Bayesian inference. CLIPScope uses a histogram of prior instance occurrences to calculate the class likelihood $p_0$ of instances being classified into class $i^*$. The histogram is updated after every inference step of the input by incrementing occurrences of $i^*$. The final OOD confidence score $p=\frac{p_1p_2}{p_0}$, thus is determined by the posterior update rule.

A key innovation of CLIPScope is it leverages the posterior information, particularly through a histogram of prior instance occurrences. This strategy significantly diverges from previous methods, which underutilized posterior information for improving OOD detection. Typically, for certain OOD samples, despite their anomalous nature, might be assigned high confidence scores by existing methods, making them difficult to detect. By incorporating posterior information, our approach greatly enhances the detection of these hard-to-distinguish OOD samples in subsequent analyses. Particularly, if hard-to-distinguish OOD samples initially bypass detection, they will contribute to an increase in the corresponding histogram bar, denoted as $c_{i^*}$. Since our overall confidence score is formulated as $p=\frac{p_1p_2}{p_0}$ with $p_0=\frac{c_{i^*}}{\sum_i c_i}$, an increment in $c_{i^*}$ results in a proportional decrement in the overall confidence score $p$. This dynamic ensures that the confidence scores assigned to these challenging OOD samples gradually decrease over time, leading to their eventual detection.  $p=\frac{p_1p_2}{p_0}$ can be interpreted as a posterior update using Bayesian inference, where $p_1p_2$ represents the joint confidence that an input instance is an ID sample and belongs to class $i^*$, and $p_0$ is the marginal likelihood, enhancing the effectiveness of OOD detection in complex and evolving data environments. It is important to mention that instead of depending on the full test samples, the calculation of $p_0$ for a particular sample relies on only the samples prior to the moment. Such a setting ensures CLIPScope to be realistic in real-world applications. 

Our contributions are manifold and can be summarized as: 1) we introduce CLIPScope, a novel zero-shot OOD detection method that leverages the prior instances to perform posterior updates, thereby enhancing the efficacy of OOD sample detection; 2) we interpret CLIPScope as a Bayesian inference problem, providing a theoretical foundation that aligns with probabilistic reasoning; 3) we propose a novel criterion for the selection of OOD labels, which contributes to the improved accuracy of our method by maximizing coverage of OOD samples; and 4) we conduct comprehensive comparisons with prior work and ablation studies.

The remainder of this paper is organized as follows: Sec.~\ref{sec:literature} delves into the background of OOD detection and CLIP. In Sec.~\ref{sec:bayesclip}, we introduce CLIPScope and provide the motivation. Sec.~\ref{sec:experiment} presents a series of comparisons and ablation studies. Sec.~\ref{sec:discussion} provides a further discussion of our approach.  Finally, Sec.~\ref{sec:conclusion} offers a conclusion to the paper.

\section{Background}
\label{sec:literature}

{\bf OOD Detection}: Classic OOD detectors have  included methods such as one-class SVM \cite{SPSSW01}, decision-tree algorithms \cite{CDGL99}, and one-class nearest neighbor approaches \cite{T02}. In more recent developments, some studies have focused on training neural networks to serve as OOD score functions. Notable examples include Deep SVDD \cite{RVGDSBMK18}, OCGAN \cite{PNX19}, GradCon \cite{KPTA20}, and Deep SAD \cite{LRN20}. Additionally, other research efforts have utilized neural networks for extracting feature embeddings, subsequently applying these embeddings to enhance OOD score functions. These post-hoc methods  \cite{QLKRM22,RCBH21,SSBRR21} do not change model parameters and use confidence scoring based on predicted softmax probabilities \cite{DK17}, energy functions~\cite{LBBH98, LWOL20, SGL21, sun2022dice}, or feature space distance\cite{LLLS18}. With the rising popularity of multimodal vision-language models, they \cite{ELRS22,MCGSLL22,WLYL23,NegLabel,SG23} have also been leveraged for OOD detection.  This recent trend signifies an evolution in the field, integrating vision-language processing capabilities to improve OOD detection.

{\bf CLIP and Zero-Shot Classification}: CLIP \cite{RKH21} is a vision-language model, trained on a vast dataset comprising approximately 400 million text-image pairs. The model is structured with two core components: a text encoder and an image encoder. For zero-shot classification tasks, CLIP employs its image encoder to extract embeddings from a given image. Concurrently, it utilizes the text encoder to generate embeddings for candidate class labels. The model then calculates the cosine similarities between the image embedding and each of the text embeddings.  CLIP   selects the ID class with the highest cosine similarity as the most likely classification for the input image. This zero-shot classification mechanism   also extends to other CLIP-based tasks. These include zero-shot object detection and zero-shot OOD detection, demonstrating CLIP's versatility and broad applicability across various domains.

\section{CLIPScope}
\label{sec:bayesclip}

%\subsection{Problem Formulation}
{\bf Problem Formulation}: Given a threshold $\gamma$, our goal  is to design a   score function $g$ to determine if an input $x$ is ID or OOD. Specifically, $x$ is considered as ID if and only if $g(x)\ge \gamma$, i.e., $x \in \text{ID if }  g(x)\ge \gamma; \text{ otherwise } x \in \text{OOD}.$

{\bf Ability}: We assume the availability of the CLIP $f$, the ID  labels $\mathcal{Y}=\{y_1, y_2, ..., y_N\}$ with $N$ representing the number of ID labels, and text from a large lexical database, such as WordNet. However, we do not have access to training ID images, ground-truth OOD labels, or prior knowledge about the distributions of ID or OOD samples in our approach.

{\bf Notation}: Let $f(x, \mathcal{Y})=y_i$ represent the process where the CLIP model $f$ receives an image $x$ and a set of ID labels $\mathcal{Y}=\{y_1, y_2, ..., y_N\}$ as inputs for the zero-shot classification. Among all the ID labels, CLIP identifies $y_i$ as the label that most closely matches  the ground-truth class of   $x$.

\subsection{Bayes' Theorem and  Intuition of CLIPScope}
%Our approach is motivated by the following lemma.
\begin{lemma}
With Bayes rule,  the class likelihood of OOD samples is proportional to the global class likelihood, i.e., $\mathbb{P}\left (f(x,\mathcal{Y})=y_i ~|~ x\in\text{OOD} \right ) \propto  \mathbb{P}(f(x,\mathcal{Y})=y_i ).$ 
\label{thm:likelihood}
\end{lemma}

{\bf Lemma}~\ref{thm:likelihood} suggests that if certain ID classes are more frequently chosen for classification during inference than others, then OOD samples are also more likely to be classified into these same ID classes. Without additional information, it's reasonable to expect these classes will continue to attract more OOD samples incorrectly as ID samples in the future. To improve future OOD sample rejection, we propose adapting the current OOD score function $g$ to make it harder for samples classified into these higher-risk classes to achieve high confidence scores. We address this by introducing a Bayesian inference-aided confidence scorer that adjusts the likelihood of a sample being OOD based on its classification into specific ID classes, offering a more nuanced and effective approach for OOD detection.

{\bf Scoring with Bayesian Inference}: Consider a random variable $x$ that represents an instance, which can be either ID or OOD. At a specific moment $t$, $x$ becomes a concrete input $x_t$, for which  we do a posterior update as follows:
\begin{align}
     \mathbb{P}&\left (x_t\in\text{ID}~|~ f(x_t,\mathcal{Y})=y_i \right ) = \\
     \nonumber
     &\frac{\mathbb{P}\left (f(x_t,\mathcal{Y})=y_i ~|~ x_t\in\text{ID} \right ) \mathbb{P}(x_t \in\text{ID}) }{ \mathbb{P}(f(x_t,\mathcal{Y})=y_i )} = \\
     \nonumber
     &\frac{\mathbb{P}\left (f(x_t,\mathcal{Y})=y_i ~|~ x_t\in\text{ID} \right ) \mathbb{P}(x_t \in\text{ID}) }{ \mathbb{P}(f(x,\mathcal{Y})=y_i )} \times C(x_t).
\end{align} Here, $C(x_t) =  \frac{\mathbb{P}(f(x,\mathcal{Y})=y_i )}{\mathbb{P}(f(x_t,\mathcal{Y})=y_i )}$ which accounts for the inherent unpredictability of the sequence of incoming samples. This adjustment ensures a consistent and fair evaluation framework across different instances, crucial for developing generalized and reliable OOD detection.

{\bf Prior Estimation}: The prior, $\mathbb{P}(x_t \in\text{ID})$, is computed based on similarities between image embedding and text embedding of both ID and OOD labels~\cite{NegLabel, ELRS22}. 
%We provide more details in Sec.~\ref{subsec:clipscope}.

{\bf  Likelihood Estimation}: We estimate the likelihood, $\mathbb{P}\left (f(x_t,\mathcal{Y})=y_i ~|~ x_t\in\text{ID} \right )$ based on  similarities between image embedding and text embedding of only ID labels.  

%We provide more details in Sec.~\ref{subsec:clipscope}.

{\bf Marginal Likelihood Estimation}: The marginal likelihood,  $\mathbb{P}(f(x,\mathcal{Y})=y_i )$ is updated based on prior sample predictions by maintaining a histogram. \textcolor{\colorname}{ We update the frequency of the predicted ID labels.} Consequently, $\mathbb{P}(f(x,\mathcal{Y})=y_i )$ is calculated by dividing the number of samples for label $y_i$ by the total samples. Thus, our confidence score function for $x_t$ is formulated as
\begin{align}
    g(x_t) = \frac{\mathbb{\hat P}\left (f(x_t,\mathcal{Y})=y_i ~|~ x_t\in\text{ID} \right ) \mathbb{\hat P}(x_t \in\text{ID}) }{ \mathbb{\hat P}(f(x,\mathcal{Y})=y_i )}.
    \label{eq:confidence}
\end{align} This formulation is distinct from previous zero-shot methods, as it incorporates the class likelihood, $ \mathbb{\hat P}(f(x,\mathcal{Y})=y_i )$, into the confidence score. This inclusion of class likelihood adjusts the confidence score in a manner that makes it challenging for samples classified into classes with high likelihood to achieve high confidence scores.

Consider an OOD instance $x_t$ with $g(x_t)>\gamma$ that is initially misclassified as an ID sample. As $x_t$ reappears, the marginal likelihood $\mathbb{\hat P}(f(x,\mathcal{Y})=y_i )$ associated with it incrementally increases. Since the prior and likelihood components of $g(x_t)$ remain constant for the same $x_t$, the increase in the marginal likelihood results in a gradual decrease of the confidence score $g(x_t)$, eventually causing it to fall below the threshold $\gamma$. While it is possible for $x_t$ to maintain a high confidence score through infrequent appearances, this sporadic occurrence inherently limits the negative impact of such hard-to-distinguish OOD samples. In essence, the confidence score function $g(x_t)$ in \eqref{eq:confidence} effectively mitigates the adverse effects of hard-to-distinguish OOD samples by either making them easier to identify over time or reducing their frequency of occurrence. 

\subsection{CLIPScope}
\label{subsec:clipscope}
The algorithm of CLIPScope, as detailed in Alg.~\ref{alg:bayesclip}, is structured into three main steps: OOD label mining (covered in lines 7-15), prior estimation (lines 18-21), and posterior updating (lines 22-26). Implementing CLIPScope requires only the ID labels $\mathcal{Y}$ with $N=|\mathcal{Y}|$, CLIP model (the text encoder $f^t$ and image encoder $f^i$), a user-defined threshold $\gamma$, and a histogram $\{c_i\}_{i=1}^N$ initialized such that each bin starts with a count of one, expressed as  $c_i=1$ $\forall i$.  

\begin{algorithm}[ht!]
\caption{CLIPScope}
\begin{algorithmic}[1]
\STATE \textbf{Input}: Text encoder $f^t$ and image encoder $f^i$ (CLIP), ID labels $\mathcal{Y}$, out/in-distribution threshold $\gamma$, \& class likelihood histogram $\{c_i\}_{i=1}^N$ initialized with $c_i=1$ for $i=1, 2,..., N$, dataset $\mathcal{X}$ where input image $x \in \mathcal{X}$
\STATE \textcolor{green!40!black}{/* Extract text embeddings of ID labels */}
\FOR{ $y_i \in \mathcal{Y}$}  
\STATE prompt = a photo of a $<y_i>$ 
\STATE $e_i = f^t(\text{prompt})/||f^t(\text{prompt})||_{L_2}$ 
\ENDFOR
\STATE \textcolor{green!40!black}{/* OOD label mining */}
\STATE $\mathcal{Y}^c = \{ \text{nouns} \in \text{WordNet}\}$ \textcolor{green!40!black}{ \/// \textit{candidate OOD labels}}
\FOR {$y_i^c \in \mathcal{Y}^c$}
\STATE prompt = a photo of a $<y_i^c>$
\STATE $\bar e_i = f^t(\text{prompt})/||f^t(\text{prompt})||_{L_2}$
\STATE $d_i$ = percentile$_\eta(\{-\text{sim}(\bar e_i, e_j\}_{j=1}^N)\}$ \textcolor{green!40!black}{\/// {\it distance to $\mathcal{Y}$}}
\ENDFOR
\STATE $\mathcal{Y}^-=\text{TopK}(\{d_i\}_{i=1}^{|\mathcal{Y}^c|}, \mathcal{Y}^c, M) $ 
\STATE $\mathcal{Y}^- = \mathcal{Y}^-\cup \text{TopK}(\{-d_i\}_{i=1}^{|\mathcal{Y}^c|}, \mathcal{Y}^c, M)$
\STATE \textcolor{green!40!black}{/* Bayesian OOD inference */}
\FORALL{$x \in \mathcal{X}$}
\STATE $h=f^i(x)/||f^i(x)||_{L_2}$
%\STATE \#\#\# Prior Estimate \#\#\# %({\it The class likelihood IS NOT taken into account})
\STATE $p_1 = \max_i \frac{e^{\text{sim}(h, e_i)/\tau}}{\sum_{j=1}^N e^{\text{sim}(h, e_j)/\tau}}$
\STATE $p_2 = \frac{\sum_{j=1}^N e^{\text{sim}(h, e_j)/\tau}}{\sum_{j=1}^N e^{\text{sim}(h, e_j)/\tau}+\sum_{j=1}^{2M} e^{\text{sim}(h, \bar e_j)/\tau}}$
\STATE $p = p_1p_2$
\STATE \textcolor{green!40!black}{/* Posterior update with class likelihoods */} %({\it The class likelihood IS taken into account})
\STATE $i^*=\argmax_i \text{sim}(h, e_i)~~~~~~$ \textcolor{green!40!black}{\/// {\it nearest ID label $y_{i^{*}}$}}
\STATE $p_0= \frac{c_{i^*}}{\sum_{i=1}^N c_i}~~$ \textcolor{green!40!black}{\/// {\it class likelihood of $y_{i^*}$ }}
\STATE $p = p/p_0\quad~~~~$   \textcolor{green!40!black}{\/// {\it adapting the confidence score}}
\STATE $c_{i^*} = c_{i^*}+1 \quad$ \textcolor{green!40!black}{\/// {\it update the histogram bar}}
\IF{$p\ge \gamma$}
\STATE $x$ is ID
\ELSE
\STATE $x$ is OOD
\ENDIF
\ENDFOR
\end{algorithmic}
\label{alg:bayesclip}
\end{algorithm}

{\bf OOD Label Mining}: This stage identifies OOD labels that are most distant from or closest to the ID labels in the embedding space  to ensure maximal coverage of potential OOD samples. CLIPScope starts by gathering a set of nouns from open-world sources (e.g., WordNet\footnote{Other similar databases can also be employed.}) as potential OOD labels, denoted as $\mathcal{Y}^c$ with a cardinality of $|\mathcal{Y}^c|$. 

The subsequent step involves extracting text embeddings for the OOD labels. A commonly used prompt format for this task is ``a photo of a $<>$'', with the blank filled by specific labels. For instance, if the label is $\{\text{cat}\}$, the generated prompts would be ``a photo of a cat''\footnote{The prompt style   can be adapted as needed.}.  CLIPScope inputs each of these prompts into the text encoder $f^t$ of the CLIP model to obtain the corresponding text embeddings. These embeddings are then normalized using the $L_2$ norm to focus solely on their directional attributes. 

To determine the distance between OOD and ID labels, our work employs a percentile distance approach, similar to the one used in NegMining \cite{NegLabel}. Specifically, for each candidate OOD label, we calculate the negative cosine similarities (denoted as sim) between its text embedding and the embeddings of all ID labels. This results in a total of N negative cosine similarities for each OOD label. The distance for each OOD label to the ID labels is then defined as the 100$\eta$-th percentile of these negative cosine similarities. In our approach, we set $\eta$ to 0.05, aligning with the practice in NegMining. This choice of $\eta=0.05$, as opposed to $\eta=0$ which represents the minimum distance, enhances robustness against outlier ID labels. After computing these distances for all OOD labels, we select those whose distances rank in the top M largest or top 
M smallest, resulting in a total of $2M$ OOD labels, defined as:
\begin{align}
\nonumber
    \mathcal{Y}^- = \text{TopK}(\{d_i\}_{i=1}^{|\mathcal{Y}^c|}, \mathcal{Y}^c, M) \cup  \text{TopK}(\{-d_i\}_{i=1}^{|\mathcal{Y}^c|}, \mathcal{Y}^c, M).
\end{align}
The  goal of deriving these OOD labels, $\mathcal{Y}^-$, is to facilitate the approximation of $\mathbb{\hat P}(x_t \in\text{ID})$ as specified in \eqref{eq:confidence}. 

{\bf Prior Estimate}: This stage computes the confidence score of the input image with only prior knowledge, which includes the ID labels $\mathcal{Y}$ and the OOD labels $\mathcal{Y}^-$. Denote $p_1 =\max_i \mathbb{\hat P}\left (f(x_t,\mathcal{Y})=y_i ~|~ x_t\in\text{ID} \right )$ as the confidence level of the zero-shot classification with only the ID labels. We begin by scaling the cosine similarities between the image embedding $h$ of the input instance and the text embeddings $e_j$  of each ID label $y_j$ by a factor $\tau$. Following this, we apply the softmax function to these scaled similarities to compute $p_1$. The equation for this computation is:
\begin{align}
    p_1 = \max_i \frac{e^{\text{sim}(h, e_i)/\tau}}{\sum_{j=1}^N e^{\text{sim}(h, e_j)/\tau}}.
\end{align}

Denote $p_2= \mathbb{\hat P}(x_t \in\text{ID})$ as the confidence level that an input image belongs to ID classes, distinct from the OOD classes, as defined in \eqref{eq:confidence}.   Unlike $p_1$, which is concerned with specific ID classes, the calculation of $p_2$ is broader, encompassing the likelihood of the image belonging to any ID class rather than specific OOD classes. To calculate $p_2$, we scale the cosine similarities between the image embedding $h$ and the text embeddings of both ID and OOD labels by $\tau$. Subsequently, these scaled similarities are processed through the softmax function. $p_2$ is obtained by summing the softmax values corresponding to all ID labels. The formula for this calculation is as follows:
\begin{align}
    p_2 = \frac{\sum_{j=1}^N e^{\text{sim}(h, e_j)/\tau}}{\sum_{j=1}^N e^{\text{sim}(h, e_j)/\tau}+\sum_{j=1}^{2M} e^{\text{sim}(h, \bar e_j)/\tau}}
\end{align} where $\bar e_j$ is the embedding of the OOD label with index $j$. This approach differs from the one used in NegLabel \cite{NegLabel} in that it does not rely on any grouping strategy. This simplifies the preprocessing steps, making our method more straightforward and efficient.

The overall OOD confidence score $p$ is the product of $p_1$ and $p_2$. This method of combining two components is also seen in the CLIPN approach \cite{WLYL23}. In CLIPN, the first component is identical to our $p_1$. However, the second component in CLIPN, $1-p^{no}$, is derived from the agreement level of their negative encoder with the CLIP model. A notable distinction between CLIPScope and CLIPN lies in the resources required. CLIPN necessitates an additional dataset containing image-text pairs for its operation. In contrast, CLIPScope relies primarily on WordNet, which provides only textual information. This difference makes our CLIPScope approach less demanding in terms of computational resources and memory usage. 

{\bf Posterior Update}: This step refines the confidence score by incorporating posterior knowledge during the inference stage.  Previous methods have not fully considered the behavior of the CLIP model across all historical instances, a factor we identify as critical for enhancing detection accuracy. CLIPScope capitalizes on this insight by analyzing the observed behavior of CLIP on historical instances, using this data to fine-tune the confidence score. This process allows for more accurate  OOD detection, as it adapts to the model's performance over time and incorporates this evolving understanding into the assessment of new instances.

Initially, CLIPScope identifies the specific ID class $i^*$ that is most closely aligned with the input image in terms of the cosine similarity between their respective embeddings, 
\begin{align}
    i^*=\argmax_i \text{sim}(h, e_i).
\end{align}
CLIPScope then proceeds to compute the likelihood $p_0$. This likelihood represents the probability of the ID class $i^*$ being the nearest to the input images based on historical instances. To achieve this, CLIPScope utilizes a histogram that tallies how frequently each ID class has historically been identified as the nearest to input images. Initially, this histogram is set with a count of one in all bins, reflecting the lack of prior posterior knowledge and implying an equal likelihood for all ID classes. As the system processes more input images, it accumulates posterior knowledge about the behavior of the CLIP model across these historical instances. The likelihood $p_0$ for the ID class  $i^*$ being the nearest to input images is approximated by the ratio of the count $c_{i^*}$ (the number of times class $i^*$ has been deemed the nearest) to the total count $\sum_{i=1}^N c_i$ of all historical instances, i.e., $ p_0= \frac{c_{i^*}}{\sum_{i=1}^N c_i}.$
The confidence score $p$ is divided by $p_0$, i.e., 
\begin{align}
    p  = \frac{p_1p_2}{\frac{c_{i^*}}{\sum_{i=1}^N c_i}} = \frac{p_1p_2}{p_0}.
    \label{eq:final}
\end{align} 
\eqref{eq:final} incorporates the evolving understanding of the CLIP model's behavior, making the CLIPScope more adaptive and accurate in its assessments over time. The histogram is updated to reflect the latest input instance's classification. This is achieved by incrementing the count for the identified nearest ID class $i^*$ by one, i.e., $c_{i^*}=c_{i^*}+1$.  CLIPScope with \eqref{eq:final} makes it challenging for instances (particularly hard-to-distinguish OOD instances) classified into classes with high likelihood to receive high OOD scores.  

%{\bf Detection}: CLIPScope compares $p$ with the threshold $\gamma$. If $p\ge \gamma$, the input is considered as ID. Otherwise, it is considered as OOD. By incorporating $p_0$ into the confidence score calculation, CLIPScope inclines to reject more instances that are classified into classes with high likelihood.

\begin{table*}[ht]
\centering

\caption{Performance (\%) of CLIPScope (ours) and other methods. The utilized CLIP model is ViT-B/16. The ID dataset is ImageNet-1k.  $\uparrow$ denotes that a higher value implies a better performance, whereas $\downarrow$ signifies that a lower value represents a better performance. }
\label{tab:comparison}
\resizebox{\textwidth}{!}{%
\begin{tabular}{@{}lcccccccccc@{}}
\toprule
OOD Dataset & \multicolumn{2}{c}{iNaturalist} & \multicolumn{2}{c}{SUN} & \multicolumn{2}{c}{Places} & \multicolumn{2}{c}{Textures} & \multicolumn{2}{c}{Average} \\ 
\cmidrule(r){2-11}
Metric & AUROC$\uparrow$ & FPR95$\downarrow$ & AUROC$\uparrow$ & FPR95$\downarrow$ & AUROC$\uparrow$ & FPR95$\downarrow$ & AUROC$\uparrow$ & FPR95$\downarrow$ & AUROC$\uparrow$ & FPR95$\downarrow$ \\ 
\addlinespace
\multicolumn{11}{c}{Methods That Require Training} \\
%\hline 
MSP & 87.44 & 58.36 & 79.73 & 73.72 & 79.67 & 74.41 & 79.69 & 71.93 & 81.633 & 69.605 \\
ODIN & 94.65 & 30.22 & 87.17 & 54.04 & 85.54 & 55.06 & 87.85 & 51.67 & 88.803 & 47.748 \\
GradNorm & 72.56 & 81.50 & 72.86 & 82.00 & 73.70 & 80.41 & 70.26 & 79.36 & 72.345 & 80.818 \\
ViM & 93.16 & 32.19 & 87.19 & 54.01 & 83.75 & 60.67 & 87.18 & 53.94 & 87.820 & 50.203 \\
KNN & 94.52 & 29.17 & 92.67 & 35.62 & 91.02 & 39.61 & 85.67 & 64.35 & 90.970 & 42.188 \\
VOS & 94.62 & 28.99 & 92.57 & 36.88 & 91.23 & 38.39 & 86.33 & 61.02 & 91.188 & 41.320 \\
NPOS & 96.19 & 16.58 & 90.44 & 43.77 & 89.44 & 45.27 & 88.80 & 46.12 & 91.218 & 37.935 \\
\addlinespace
\multicolumn{11}{c}{Zero-Shot (No Training Required)} \\
Mahalanobis & 55.89 & 99.33 & 59.94 & 99.41 & 65.96 & 98.54 & 64.23 & 98.46 & 61.505 & 98.935 \\
Energy & 85.09 & 81.08 & 84.24 & 79.02 & 83.38 & 75.08 & 65.56 & 93.65 & 79.568 & 82.208 \\
ZOC & 86.09 & 87.30 & 81.20 & 81.51 & 83.39 & 73.06 & 76.46 & 98.90 & 81.785 & 85.193 \\
MCM & 94.59 & 32.20 & 92.25 & 38.80 & 90.31 & 46.20 & 86.12 & 58.50 & 90.818 & 43.925 \\
CLIPN & 95.27 & 23.94 & 93.93 & 26.17 & 92.28 & 33.45 & 90.93 & 40.83 & 93.103 & 31.098 \\
NegLabel & {99.49} & {1.91} & 95.49 & 20.53 & 91.64 & 35.59 & 90.22 & 43.56 & 94.210 & 25.398 \\
\hdashline
Ours (Random) & {\bf 99.61 }  &	{\bf 1.29  } &	{\bf 96.77  } &	{\bf 15.56  } &	{\bf 93.54  } &	{\bf 28.45  } &	{\bf 91.41  } &	{\bf 38.37  } &	{\bf 95.301  } &	{\bf 20.883  }  \\
%\hdashline
Ours (Forward) & 99.64 &	1.42 &	96.1 &	17.6 &	92.73 &	30.16 &	89.76 &	42.35 &	94.558 &	22.883 \\
Ours (Reverse) & 99.60 &	1.28 &	97.34 &	13.52 &	94.20 &	26.32 &	93.04 &	34.41 &	96.045 &	18.883\\
\bottomrule
\end{tabular}%
}
\vspace{-0.3cm}
\end{table*}

\section{Experimental Results}
\label{sec:experiment}
%\subsection{Setup}
{\bf Datasets and  Metrics}: We employ ImageNet-1K \cite{HGL21} as the primary ID dataset. For OOD datasets, we utilize iNaturalist \cite{VM18}, SUN \cite{XH10}, Places \cite{ZL17}, and Textures \cite{CMKMV14}. These datasets are widely used in zero-shot OOD detection research for method evaluation. They encompass large-scale visual images across diverse scenes and semantics, offering a comprehensive testing ground. The OOD datasets are carefully curated to ensure no class overlap with the ID dataset, thus maintaining distinct separations between ID and OOD categories. Our method's efficacy is evaluated using standard metrics prevalent in the field: the area under the receiver operating characteristic curve (AUROC) and the false positive rate at a 95\% true positive rate (FPR95). A higher AUROC indicates greater detection accuracy, while a lower FPR95 denotes superior performance.

{\bf Implementation Details}: For our zero-shot OOD detection, we utilize the CLIP model \cite{RKH21}, specifically the CLIP-B/16 variant, which is a common choice in related studies. During the OOD label mining phase, WordNet serves as our chosen corpus, with $M=5000$ set as the parameter, leading to a selection of $2M=10000$ words as OOD labels. Consistent with the methodology in NegLabel, we set the percentile distance to $\eta=0.05$ and the scaled temperature to $\tau=0.01$.  Our experiments are conducted using an NVIDIA Quadro RTX 5000 GPU.

%\subsection{Comparison Results}

{\bf Baseline Methods}: The compared zero-shot OOD detection methods include Mahalanobis distance \cite{LLLS18}, energy score \cite{LWOL20}, ZOC \cite{ELRS22}, MCM \cite{MCGSLL22}, CLIPN \cite{WLYL23}, and NegLabel \cite{NegLabel}. We also compare our approach with  methods that train or fine-tune OOD detection models with ID data, including MSP \cite{DK17}, ODIN \cite{LLS18}, GradNorm \cite{HGL21}, ViM \cite{WLFZ22}, KNN \cite{SMZL22}, VOS \cite{XZMS22}, and NPOS \cite{TDZL22}.  MSP uses the highest softmax output as a confidence measure. The KNN approach adopts a contrastive learning framework  tailored for OOD detection. ODIN enhances detection by merging temperature scaling with a gradient-based input pre-processing. GradNorm differentiates ID from OOD data by comparing the magnitudes of gradient norms, positing that ID data typically results in larger gradients. ViM introduces an additional logit in the output layer, which signifies the confidence level of an input being OOD after applying softmax. VOS generates virtual OOD samples for training effective detectors.  NPOS fine-tunes CLIP's image encoder with synthetic OOD data to enhance detection capabilities. 

{\bf Sequence Order}: The performance of CLIPScope is influenced by the sequence order in which ID and OOD samples are processed, as $p_0$ is calculated based on historical instances. We consider three different scenarios: forwarding, reversing, and random orders.  In forwarding order, ID instances are processed before all OOD instances.  In reversing order,  ID instances follow after all OOD instances. In random order, ID and OOD instances are intermixed and appear in a random sequence. Note that in practical applications, the random order is more likely to happen without knowing the positions of OOD inputs. Therefore, we assessed CLIPScope across various random sequences by conducting five independent trials and calculating the average. However, we also provide the results on forwarding and reversing orders for certain cases to show the potential range of our approach's performance. 

{\bf Large-Scale Datasets}:  The performance of CLIPScope in the random order scenario is presented in Table~\ref{tab:comparison}. CLIPScope shows the highest AUROC and lowest  FPR95 for all cases.  Table~\ref{tab:comparison} also shows the performance of CLIPScope in the forwarding and reversing order scenarios. CLIPScope demonstrates its highest performance in the reversing order, while the lowest performance is observed in the forwarding order. The reason behind this trend is that when calculating confidence scores for OOD samples in the forwarding order, the histogram $c_i$ has been initialized by the distribution of ID samples. Therefore, the deviation of  $c_i$ from the actual OOD distribution is the largest among the three cases, leading to the worst performance.  In contrast, the reversing order has the smallest distributional deviation shift, leading to the best performance.  The distributional deviation shift of the random order case is in the middle and so is the performance.  Notably, even under the forwarding order scenario, CLIPScope surpasses existing approaches across most cases, demonstrating its efficacy. 

\begin{figure*}
    \centering
    \includegraphics[width=0.24\linewidth]{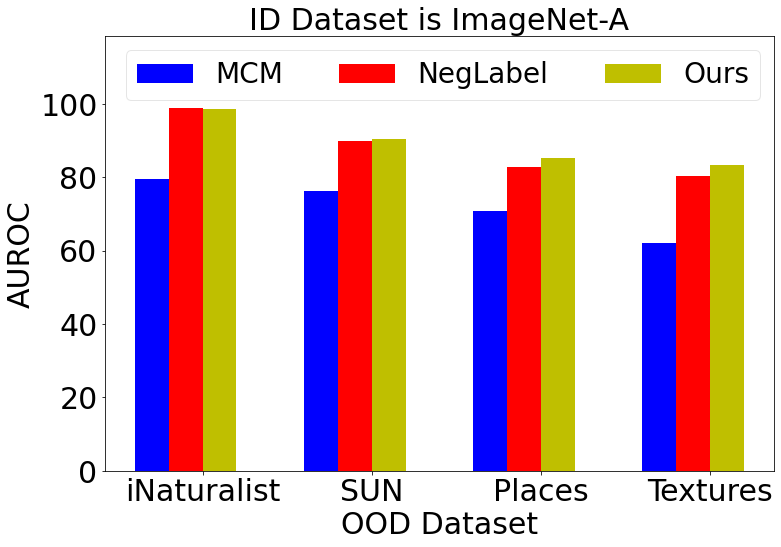}
    \includegraphics[width=0.24\linewidth]{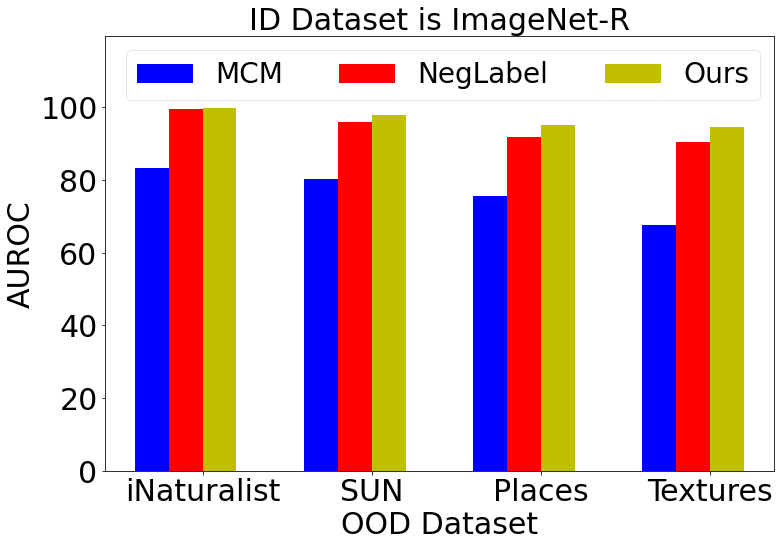}
    \includegraphics[width=0.24\linewidth]{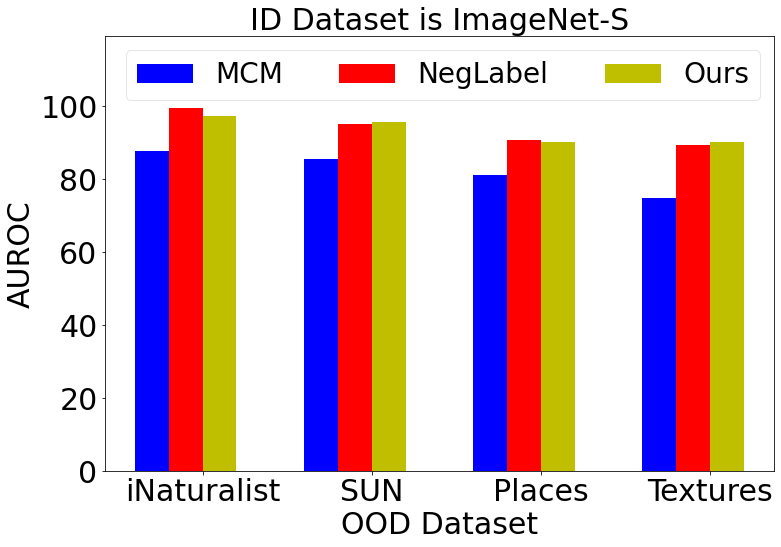}
    \includegraphics[width=0.24\linewidth]{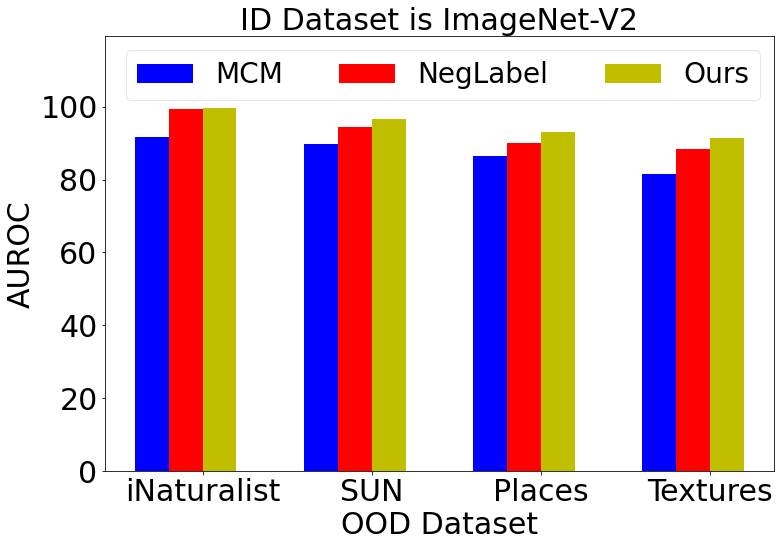}
    \caption{AUROC (\%)  on domain-shifted ID datasets. A higher AUROC implies a better performance.}
    \label{fig:shift}
    \vspace{-0.3cm}
\end{figure*}

{\bf Domain-Shifted ID Datasets}: CLIPScope demonstrates notable robustness when applied to domain-shifted ImageNet datasets. In our evaluations, we  utilized ImageNet-A \cite{HZBSS21}, ImageNet-R \cite{HBMK21}, ImageNet-Sketch \cite{WG19}, and imageNet-V2 \cite{RRSS19} as the ID datasets. The results of these tests are visually represented in Fig.\ref{fig:shift}. In these domain-shifted scenarios, CLIPScope shows comparable performance to NegLabel.  This highlights the method's capability to adapt to various data distributions effectively. In contrast, MCM exhibits a noticeable decrease in performance on these domain-shifted ID datasets. 

{\bf Contribution of $p_0$, $p_1$, and $p_2$}: We conducted an ablation analysis to assess the individual and combined effectiveness of the components prior ($p_0$), likelihood ($p_1$), and marginal likelihood ($p_2$). The results of this analysis are detailed in Table~\ref{tab:ablation}. The findings reveal that the incorporation of $p_0$ significantly enhances OOD detection accuracy, even when it is the sole component used in conjunction with either $p_1$ or $p_2$ as the numerator in the confidence score calculation. This underscores the vital role of $p_0$ in the algorithm. Furthermore, the results indicate that employing all three components in the confidence score function leads to best performance for most cases, both in terms of AUROC and FPR95. The data in the row  $g=p_1/p_0$ illustrate that CLIPScope distinguishes itself from NegLabel by remaining effective without using any negative OOD labels.

\begin{table*}
\centering
\caption{Performance (\%) of CLIPScope with various confidence scores $g$ (the top table) and OOD label space $\mathcal{Y}^-$ (the middle and bottom tables). The ID dataset is ImageNet-1k. }
\label{tab:ablation}
\resizebox{\textwidth}{!}{%
\begin{tabular}{@{}lcccccccc@{}}
\toprule
OOD Dataset & \multicolumn{2}{c}{iNaturalist} & \multicolumn{2}{c}{SUN} & \multicolumn{2}{c}{Places} & \multicolumn{2}{c}{Textures}  \\ 
\cmidrule(r){2-9}
Metric & AUROC$\uparrow$ & FPR95$\downarrow$ & AUROC$\uparrow$ & FPR95$\downarrow$ & AUROC$\uparrow$ & FPR95$\downarrow$ & AUROC$\uparrow$ & FPR95$\downarrow$  \\ 

\addlinespace
\multicolumn{9}{c}{Confidence Score $g$} \\
$g = p_1$ & 80.19 &	70.43 &	85.01 &	55.52 &	83.41 &	57.03 &	79.75 &	63.24  \\
$g = p_1/p_0$ & {\bf 97.98} &	{\bf 8.23} &	{\bf 95.79} &	{\bf 18.38} &	{\bf 91.81} & {\bf 30.68} & {\bf 92.36} &	{\bf 31.52} \\
\hdashline
$g = p_2$ & 99.31 &	2.18 &	91.13 &	43.51 &	88.07 &	53.81 &	78.32 &	73.01	 \\
$g = p_2/p_0$ & {\bf 99.54}	 & {\bf 1.58}	 & {\bf 96.33} &	{\bf 17.31} &	{\bf 92.54} &	{\bf 31.67} &	{\bf 90.41} &	{\bf 41.58} \\
\hdashline 
$g = p_1p_2$ & 99.19 &	3.15 &	91.49 &	34.65 &	88.89	 & 43.09& 	81.03& 	60.9 \\
$g = p_1p_2/p_0$ & {\bf 99.61 }  &	{\bf 1.29  } &	{\bf 96.77  } &	{\bf 15.56  } &	{\bf 93.54  } &	{\bf 28.45  } &	{\bf 91.41  } &	{\bf 38.37  }    \\

\addlinespace
\multicolumn{9}{c}{Different OOD Label Space $\mathcal{Y}^-$} \\
$M=5000$, Nearest \& Farthest &  99.61   &	 1.29   &	{\bf 96.77  } &	{\bf 15.56  } &	{\bf 93.54  } &	{\bf 28.45  } &	 91.41   &	 38.37     \\
$M=5000$, Nearest Only & {\bf 99.69} &	{\bf 1.01} &	96.56 &	15.98 &	93.06 &	30.02 &	91.18 &	39.27 \\
$M=5000$, Farthest Only & 97.39 &	12.95 &	96.41 &	17.57 &	92.91 &	29.35	& {\bf 92.08} &	{\bf 36.97}  \\

\addlinespace
\multicolumn{9}{c}{OOD Label Space $\mathcal{Y}^-$ Overlap With ID Label Space $\mathcal{Y}$} \\
$M=5000$, Nearest \& Farthest  $\cup$ $\mathcal{Y}$ & 99.61	 & 1.36 &	96.79 &	15.27 &	93.48 &	27.92 &	91.84 &	36.93 \\
$M=10000$, Nearest Only $\cup$ $\mathcal{Y}$  & 99.66 &	1.08 &	96.87 &	14.56 &	93.31 &	28.91 &	91.35 &	38.61  \\
\bottomrule
\end{tabular}%
}
\vspace{-0.5cm}
\end{table*}

{\bf Contribution of Nearest and Farthest OOD Labels}: We explored variations of the label selection process. These variations included selecting  only the nearest OOD labels and only the farthest OOD labels. The results are presented in Table~\ref{tab:ablation}. The results indicate that both farthest and nearest OOD labels help contribute to detecting OOD samples. Therefore, combining them achieves the best or second-best performance across all tested scenarios.  \textcolor{\colorname}{
As stated previously, we include labels close to the ID labels within the OOD labels to maximize coverage of the OOD label space.Table~\ref{tab:ablation} demonstrates that using both nearest and farthest OOD labels outperforms using only the farthest OOD labels in three out of four OOD datasets (iNaturalist, SUN, and Places). This is likely because the model's ability to distinguish between ID and OOD samples is enhanced by capturing a more diverse range of OOD examples.}

{\bf Noisy OOD Labels}:  Our confidence scorer contains three components: $p_0$, $p_1$, and $p_2$. Neither $p_0$ nor $p_1$ is influenced by OOD labels. As illustrated in Table~\ref{tab:ablation}, our approach is effective with the confidence score $g=p_1/p_0$, being inherently robust against the presence of noisy OOD labels that might fall within ID space. Moreover, we conducted targeted experiments by constructing OOD labels from a mix of 5000 nearest OOD, 5000 farthest OOD, and 1000 ID labels, intentionally creating conditions where OOD labels  intersect with ID labels. The result is shown in the bottom of Table~\ref{tab:ablation}.  \textcolor{\colorname}{Performance degradation is not apparent because (1) the number of OOD labels is reasonably higher than the number of ID labels. And (2) only $p_2$ is affected when ID labels are added to the OOD labels. However, the potential inaccuracy of $p_2$ is mitigated by the other two components, $p_0$ and $p_1$, since they are not affected and the performance of the scorer $p_1/p_2$ is reasonably accurate.} A similar outcome was observed with an alternative label set comprising 10000 nearest OOD and 1000 ID labels, further reinforcing our method's robustness.  {\bf Additional Evaluations} are provided in our supplementary material.

\section{Discussion}
\label{sec:discussion}

{\bf Comparisons with NegLabel}: The core innovation of our approach lies in the Bayesian OOD scorer. This scorer leverages the mined coverage-maximizing OOD labels in conjunction with historical data through the marginal likelihood term $p_0$. By dynamically adapting confidence scores in real-time, this unique feature enables our method to significantly improve OOD detection performance. As demonstrated in Table~\ref{tab:ablation}, the inclusion of $p_0$ alone contributes to a substantial increase in AUROC, highlighting the effectiveness and novelty of our approach.  While $p_2$ bears similarity to the NegLabel score, it is not identical in that it does not rely on any grouping strategy, therefore simplifying the preprocessing steps and removing need for heuristic computations.   Moreover, our method remains robust even without the $p_2$ term, as evidenced by the results in the $g=p_1/p_0$ row of Table~\ref{tab:ablation}.  The strategic adaptation enabled by $p_0$ represents a significant advancement in the field of zero-shot OOD detection by introducing real-time adaptability to previously static detection frameworks. {\bf Additional discussions} are given in our supplementary material.

\section{Conclusion}
\label{sec:conclusion}
We have presented CLIPScope to detect OOD samples in a zero-shot manner, leveraging Bayesian inference to enhance confidence scoring. It uses information from prior instances as evidence for posterior updates to effectively identify OOD samples. Furthermore, we have developed a unique OOD label mining strategy for CLIPScope scoring, enhancing OOD sample coverage by utilizing the closest and farthest WordNet labels based on their CLIP embedding proximity to ID classes. We perform extensive comparisons with prior approaches, demonstrating excellent performance, and conduct ablation studies to validate the effectiveness of CLIPScope. 

\newpage
%%%%%%%%% REFERENCES
{\small
\bibliographystyle{ieee_fullname}
\bibliography{example_paper}
}

\newpage
\onecolumn
\appendix
\section{Proof of Lemma~\ref{thm:likelihood}}
\begin{proof} 
\begin{align}
\nonumber
     \mathbb{P}\left (f(x,\mathcal{Y})=y_i ~|~ x\in\text{OOD} \right ) = \frac{\mathbb{P}\left (x\in\text{OOD} ~|~ f(x,\mathcal{Y})=y_i \right ) \mathbb{P}(f(x,\mathcal{Y})=y_i)}{\mathbb{P}(x\in\text{OOD})} 
     \propto  \mathbb{P}(f(x,\mathcal{Y})=y_i ).
\end{align}
\end{proof}

\section{Additional Evaluations}
\subsection{Hard OOD Tasks}

We evaluate CLIPScope on hard OOD tasks. The results are shown in Table~\ref{tab:HardOOD}. Our approach shows performance comparable to NegLabel.

\begin{table}[ht]
\caption{Comparisons on hard OOD tasks. In each case, ID dataset is shown in the top, whereas OOD dataset is shown in the bottom. N/A represents that the corresponding results are not provided in the original paper.}
    \centering
    \begin{tabular}{c|cc|cc}
    \hline
       & \multicolumn{2}{c|}{CLIPScope } & \multicolumn{2}{c}{NegLabel }  \\
    &  AUROC & FPR95 & AUROC & FPR95   \\
    \hline
  ImageNet-10     & \multirow{2}{*}{98.41} & \multirow{2}{*}{7} & \multirow{2}{*}{98.86} & \multirow{2}{*}{5.1}    \\
     ImageNet-20 & & & & \\
     \hdashline
      ImageNet-10     & \multirow{2}{*}{98.89} & \multirow{2}{*}{2} & \multirow{2}{*}{99.51} & \multirow{2}{*}{1.68}    \\
     ImageNet-100 & & & & \\
     \hdashline
      ImageNet-20     & \multirow{2}{*}{98.42	} & \multirow{2}{*}{6.8} & \multirow{2}{*}{98.81} & \multirow{2}{*}{4.6}    \\
     ImageNet-10 & & & & \\
     \hdashline
     ImageNet-20     & \multirow{2}{*}{97.43	} & \multirow{2}{*}{11.98} & \multirow{2}{*}{N/A} & \multirow{2}{*}{N/A}    \\
     ImageNet-100 & & & & \\
     \hdashline
     ImageNet-100     & \multirow{2}{*}{92.11} & \multirow{2}{*}{25.8} & \multirow{2}{*}{90.19} & \multirow{2}{*}{40.2}    \\
     ImageNet-10 & & & & \\
     \hdashline
     ImageNet-100     & \multirow{2}{*}{89.83} & \multirow{2}{*}{27.6} & \multirow{2}{*}{N/A} & \multirow{2}{*}{N/A}    \\
     ImageNet-20 & & & & \\
     \hline
    \end{tabular}
    \label{tab:HardOOD}
\end{table}

\subsection{FPR95 for Tests in Fig.~\ref{fig:shift}}

\begin{figure*}[ht]
    \includegraphics[width=0.24\linewidth]{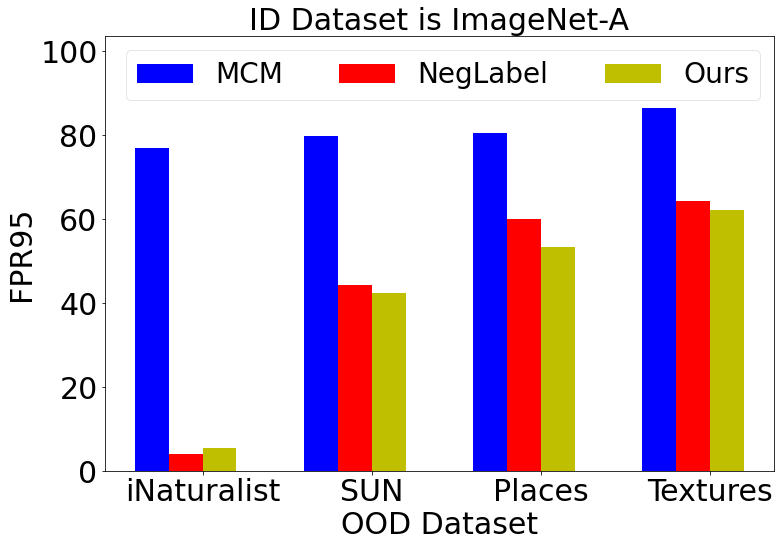}
    \includegraphics[width=0.24\linewidth]{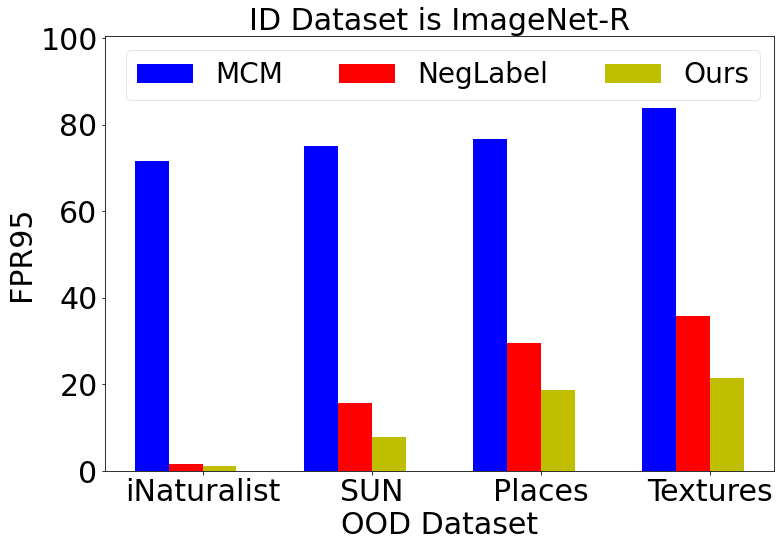}
    \includegraphics[width=0.24\linewidth]{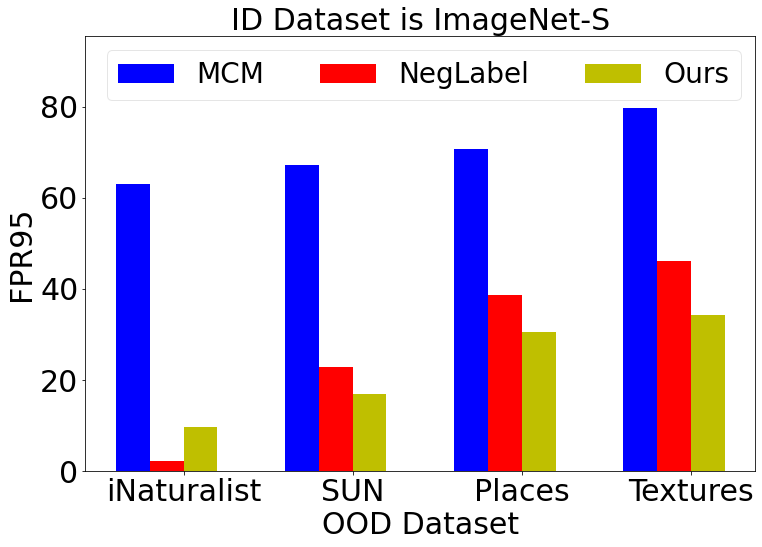}
    \includegraphics[width=0.24\linewidth]{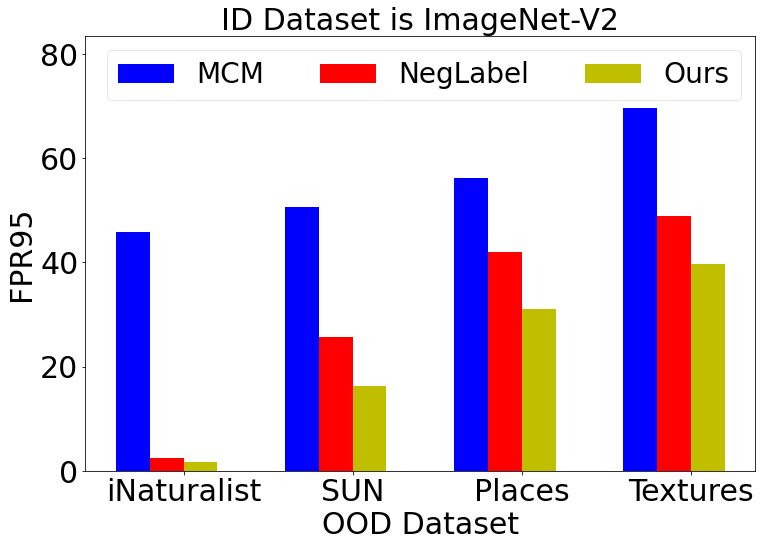}
     \caption{FPR95 (\%)  on domain-shifted ID datasets. A lower FPR95 implies a better performance.}
    \label{fig:shift2}
\end{figure*}

\subsection{Small ID Datasets}

\begin{figure}[ht]
    \centering
    \includegraphics[width=0.3\linewidth]{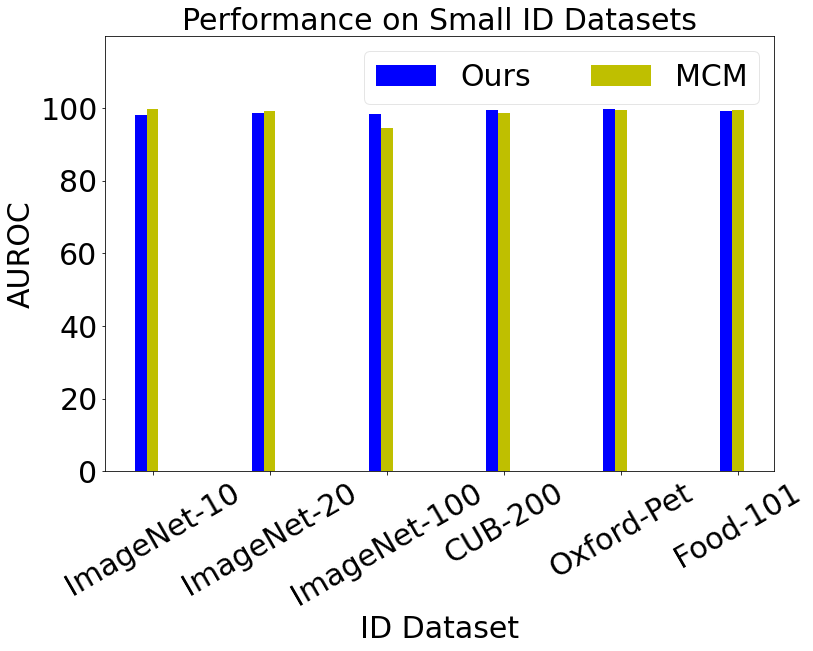}
    \includegraphics[width=0.3\linewidth]{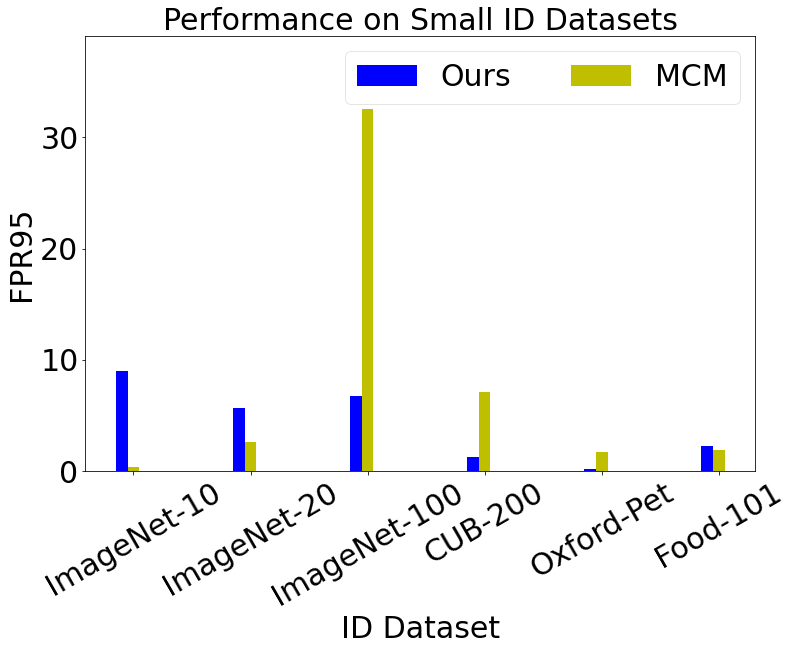}
    \caption{Performance (in \%) of CLIPScope when applied to small ID datasets. The OOD datasets include iNaturalist, SUN, Places, and Textures. The reported numbers represent average results across these four OOD datasets. }
    \label{fig:small}
    %\vspace{-0.5cm}
\end{figure}

We conducted further experiments with smaller ID datasets. The results are shown in Fig.~\ref{fig:small}. Our approach consistently maintained its effectiveness across these smaller datasets.   \textcolor{\colorname}{ Specifically, Table~\ref{tab:app_smallID} presents the FPR95 and AUROC values corresponding to Fig.~\ref{fig:small}. Our approach maintains consistently high AUROC across all small ID datasets. The lower FPR95 values observed in CUB-200, Oxford-Pet, and Food-101 can be attributed to their focus on fine-grained categories (e.g., specific bird species, pet breeds, or food types). These datasets contain highly specific and detailed visual features within each class, distinguishing them from OOD datasets. For ImageNet-10, ImageNet-20, and ImageNet-100 as ID datasets, the average FPR95 ranges from 5\% to 9\%. While still relatively low, this slight performance drop can be attributed to their diverse classes with limited images per class. } The results show the ability of our approach to deliver reliable OOD detection performance regardless of the ID dataset size.

\begin{table}[ht!]
        \centering
        \textcolor{\colorname}{
        \caption{Performance of CLIPScope on each small ID dataset. }
\label{tab:app_smallID}
\resizebox{\textwidth}{!}{%
\begin{tabular}{@{}lcccccccccc@{}}
\toprule
OOD Dataset & \multicolumn{2}{c}{iNaturalist} & \multicolumn{2}{c}{SUN} & \multicolumn{2}{c}{Places} & \multicolumn{2}{c}{Textures} & \multicolumn{2}{c}{Average} \\ 
\cmidrule(r){2-11}
Small ID Datasets & AUROC$\uparrow$ & FPR95$\downarrow$ & AUROC$\uparrow$ & FPR95$\downarrow$ & AUROC$\uparrow$ & FPR95$\downarrow$ & AUROC$\uparrow$ & FPR95$\downarrow$ & AUROC$\uparrow$ & FPR95$\downarrow$ \\ 
%\addlinespace
ImageNet-10 & 99.85 &	0.64 &	97.84 &	10.63 &	96.33 &	18.80 &	98.62 &	5.81 &	98.159 &	8.965 \\
ImageNet-20 & 99.90	 & 0.42 &	98.74 &	7.24 &	97.84 &	10.96 &	98.69 &	4.07 &	98.789 &	5.671 \\
ImageNet-100 & 99.67 &	1.28 &	98.66 &	5.65 &	97.44	 & 10.64 &	97.55 &	9.37 &	98.329 &	6.734 \\
CUB-200 & 99.78 &	0.65 &	99.68 &	0.87 &	99.23 &	2.56 &	99.67 &	1.11 &	99.589 &	1.294 \\
Oxford-Pet & 99.99 &	0.02 &	99.97 &	0.04 &	99.88 &	0.36 &	99.85 &	0.33 &	99.923 &	0.185 \\
Food-101 & 99.97 &	0.11 &	99.83 &	0.61 &	99.63 &	1.40 &	97.26 &	6.91 &	99.169 &	2.255 \\
\hline
\end{tabular}
}}
\end{table}

\subsection{Robustness Against Mining Parameters $M$ and $\eta$}
 We assessed CLIPScope across various sizes $M$ of OOD label space and percentile distances $\eta$. To mitigate the effects of randomness, we employed a reversing order. The findings are detailed in Table~\ref{tab:mining}. $M$ and $\eta$ exert only a mild influence on our approach since only $p_2$ utilizes OOD labels.

\begin{table}
\centering
\caption{Performance (\%) of CLIPScope with various $M$ (the top table) and $\eta$ (the bottom table). The ID dataset is ImageNet-1k.  }
\label{tab:mining}
\resizebox{\textwidth}{!}{%
\begin{tabular}{@{}lcccccccccc@{}}
\toprule
OOD Dataset & \multicolumn{2}{c}{iNaturalist} & \multicolumn{2}{c}{SUN} & \multicolumn{2}{c}{Places} & \multicolumn{2}{c}{Textures} & \multicolumn{2}{c}{Average}  \\ 
\cmidrule(r){2-11}
Metric & AUROC$\uparrow$ & FPR95$\downarrow$ & AUROC$\uparrow$ & FPR95$\downarrow$ & AUROC$\uparrow$ & FPR95$\downarrow$ & AUROC$\uparrow$ & FPR95$\downarrow$  & AUROC$\uparrow$ & FPR95$\downarrow$  \\ 

\addlinespace
\multicolumn{11}{c}{Different Sizes $M$ of OOD Label Space $\mathcal{Y}^-$ (Nearest \& Farthest)} \\
$M=0$ & {97.98} &	{8.23} &	{95.79} &	{18.38} &	{91.81} & {30.68} & {92.36} &	{31.52} & 94.488 &	22.204 \\
$M=50$  & 98.41 &	5.28 &	96.63 &	15.48	 & 93.03 &	27.48 &	93.55 &	27.26 &	95.405 &	18.875 \\
$M=100$ & 98.68 &	4.87 &	96.7 &	15.23	 &93.19 &	27.23 &	93.69 &	26.93 &	95.565 &	18.565 \\
$M=500$ & 99.29 &	2.27 &	97.06 &	13.55 &	93.70 &	25.57 &	93.81 &	28.58 &	95.965 &	17.493 \\
$M=1000$ & 99.45 &	1.52 &	97.12 &	13.64 &	93.85 &	25.43 &	93.65 &	30.3 &	96.018 &	17.723\\
$M=2000$ & 99.53 &	1.35 &	97.25 &	13.76 &	94.1 &	25.54	 & 93.44	 &32.09 &	96.080 &	18.185  \\
$M=5000$ & 99.60 &	1.28 &	97.34 &	13.52 &	94.20 &	26.32 &	93.04 &	34.41 &	96.045 &	18.883  \\
$M=7000$ & 99.60 &	1.21 &	97.41 &	12.91 &	94.27 &	26.14 &	92.85 &	35.12 &	96.033 &	18.845 \\
$M=10000$ & 99.60 &	1.23 &	97.47 &	12.83 &	94.3 &	25.69 &	92.91 &	34.23 &	96.070 &	18.495 \\

\addlinespace
\multicolumn{11}{c}{Different Percentile Distance $\eta$} \\
$\eta = 0.001$  & 99.49 &	1.68 &	97.10 &	12.25 &	95.12 &	21.12 &	92.90 &	31.39 &	96.153 &	16.610 \\
$\eta = 0.05$  & 99.60 &	1.28 &	97.34 &	13.52 &	94.20 &	26.32 &	93.04 &	34.41 &	96.045	 & 18.883  \\
$\eta = 0.25$  & 99.52 &	1.56 &	96.92 &	13.68 &	94.75 &	23.06 &	92.55 &	32.56 &	95.935 &	17.715 \\
$\eta = 0.5$  & 99.51 &	1.52 &	96.94 &	13.37 &	94.80	 & 22.93 &	92.56 &	32.68 &	95.953 &	17.625 \\
$\eta = 0.75$  & 99.58 &	1.3 &	97.31 &	13.51 &	94.19	 & 26.41 &	93.24 &	33.70 &	96.080 &	18.730 \\
$\eta = 0.95$  & 99.39 &	2.03 &	97.52 &	11.77 &	94.36 &	24.87 &	93.33 &	31.64 &	96.150	 & 17.578 \\
$\eta = 0.999$  & 99.27 &	2.91 &	97.57 &	11.02 &	93.97 &	25.56 &	92.95 &	32.5 &	95.940 &	17.998 \\
\bottomrule
\end{tabular}%
}
\end{table}

\subsection{Convergence}
\begin{figure*}[ht]
    \centering
    \includegraphics[width=0.24\linewidth]{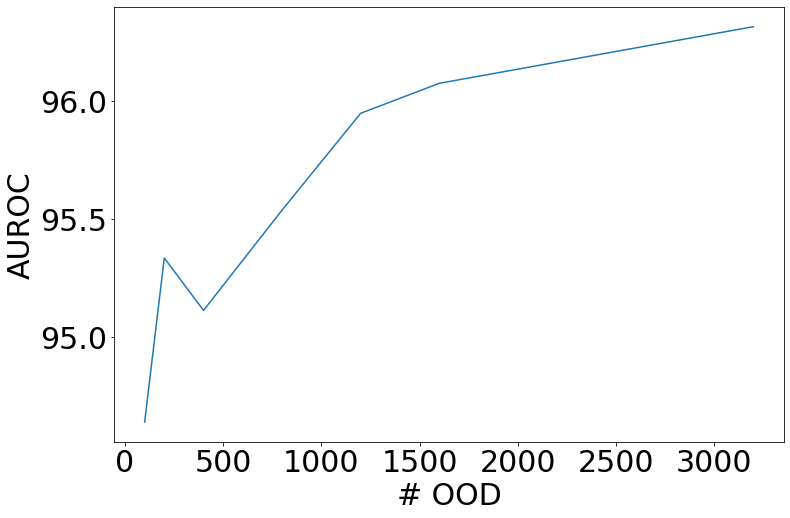}
    \includegraphics[width=0.24\linewidth]{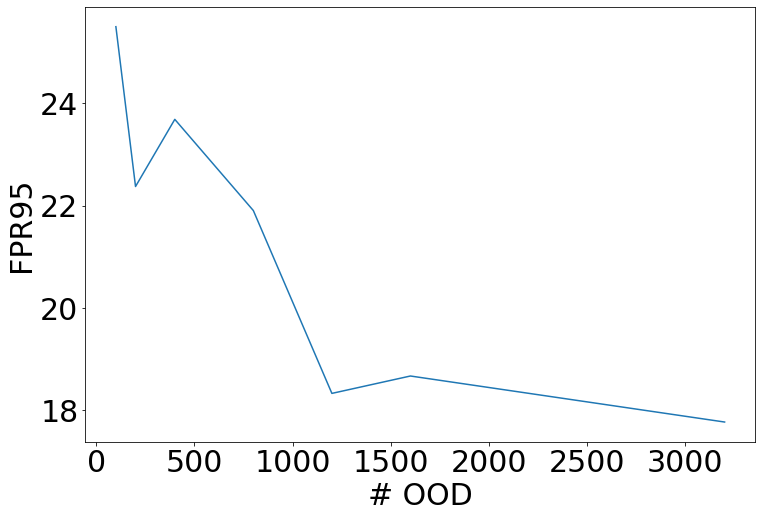}
    \includegraphics[width=0.24\linewidth]{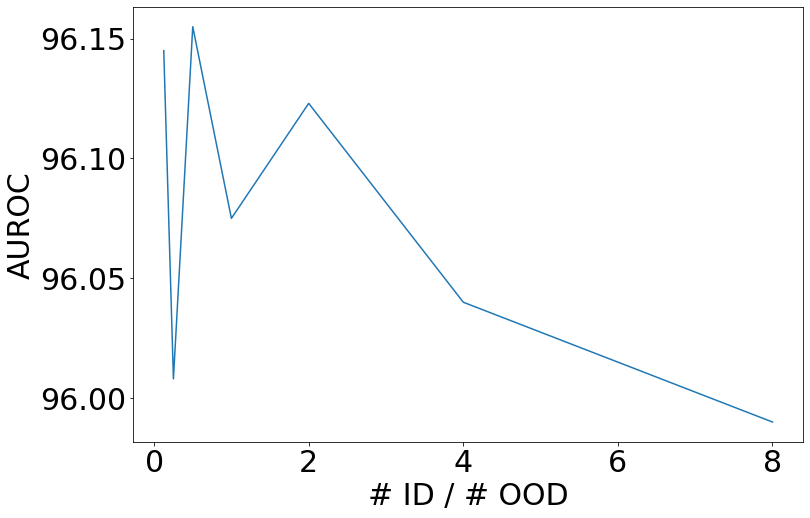}
    \includegraphics[width=0.24\linewidth]{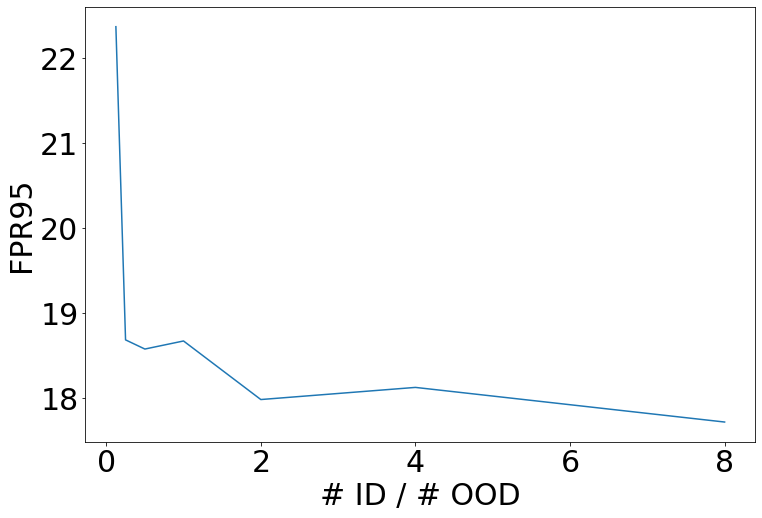}
    \caption{Performance (\%) of CLIPScope across different quantities of OOD samples (top), and varying ratios of ID to OOD samples (bottom). The ID dataset is ImageNet-1k. The OOD datasets include iNaturalist, SUN, Places, and Textures. The figures presented are the average results  from these four cases. }
    \label{fig:converge}
\end{figure*}

We conducted a series of experiments by varying the numbers of OOD samples to assess the impact on performance stability. The ratio of ID to OOD samples is maintained at 1:1. The results are shown in the first two subplots of Fig.~\ref{fig:converge} and \textcolor{\colorname}{Table~\ref{tab:app_curve}}. These numbers indicate that our approach reaches a performance plateau after processing approximately 1200 OOD samples, which is about 12\% of the total OOD samples included in our test sets. Similarly, we conducted experiments to explore how varying proportions affect our approach's efficacy. We fix the number of OOD samples at 1600 and vary the number of ID samples. The results are shown in the last two subplots of Fig.~\ref{fig:converge} and \textcolor{\colorname}{Table~\ref{tab:app_unbalance}}. \textcolor{\colorname}{ Our methodology shows performance plateaus at 2000 samples in this case. }

\begin{table}
        \centering
       \textcolor{\colorname}{
        \caption{Performance of CLIPScope on each small ID dataset. Each case contains 50\% ID samples and 50\% OOD samples. }
\label{tab:app_curve}
\resizebox{\textwidth}{!}{%
\begin{tabular}{@{}lcccccccccc@{}}
\toprule
OOD Dataset & \multicolumn{2}{c}{iNaturalist} & \multicolumn{2}{c}{SUN} & \multicolumn{2}{c}{Places} & \multicolumn{2}{c}{Textures} & \multicolumn{2}{c}{Average} \\ 
\cmidrule(r){2-11}
\# Samples & AUROC$\uparrow$ & FPR95$\downarrow$ & AUROC$\uparrow$ & FPR95$\downarrow$ & AUROC$\uparrow$ & FPR95$\downarrow$ & AUROC$\uparrow$ & FPR95$\downarrow$ & AUROC$\uparrow$ & FPR95$\downarrow$ \\ 
%\addlinespace
200 &	99.02 &	4 &	93.91 &	30 &	90.3 &	39 &	95.33 &	29 &	94.640 &	25.500 \\ 
400 &	99.35 &	3.5 &	94.9 &	26.5 &	90.9 &	38.5 &	96.19 &	21 &	95.335 &	22.375 \\ 
800 &	99.16 &	3.75 &	95.06 &	22 &	94.16 &	24.75 &	92.07 &	44.25 &	95.113 &	23.688 \\ 
1600 &	99.33 &	2.5 &	96.63 &	16.37 &	94.56 &	25.25	& 91.63 &	43.5 &	95.538 &	21.905 \\ 
2400 &	99.42 &	1.66 &	96.96 &	15.08 &	94.58 &	23 &	92.83 &	33.58 &	95.948 &	18.330 \\ 
3200 &	99.44 &	1.5 &	96.98 &	13.75 &	94.98 &	24.75	 & 92.90 &	34.68 &	96.075 &	18.670 \\ 
6400 &	99.52	 & 1.34 &	97.30 &	13 &	95.12 &	23.53 &	93.32 &	33.21 &	96.315 &	17.770 \\ 
12800 &	99.55 &	1.23 &	97.43 &	12.68 &	94.34 &	26.73 &	93.15 &	35 &	96.118 &	18.910 \\ 
\hline
\end{tabular}
}}
\end{table}

    \begin{table}
        \centering
       \textcolor{\colorname}{
        \caption{Performance of CLIPScope on various ID/OOD ratios. The number of OOD samples is fixed at 1600. }
\label{tab:app_unbalance}
\resizebox{\textwidth}{!}{%
\begin{tabular}{@{}lcccccccccc@{}}
\toprule
OOD Dataset & \multicolumn{2}{c}{iNaturalist} & \multicolumn{2}{c}{SUN} & \multicolumn{2}{c}{Places} & \multicolumn{2}{c}{Textures} & \multicolumn{2}{c}{Average} \\ 
\cmidrule(r){2-11}
\# ID / \# OOD & AUROC$\uparrow$ & FPR95$\downarrow$ & AUROC$\uparrow$ & FPR95$\downarrow$ & AUROC$\uparrow$ & FPR95$\downarrow$ & AUROC$\uparrow$ & FPR95$\downarrow$ & AUROC$\uparrow$ & FPR95$\downarrow$ \\ 
%\addlinespace
1/8 &	99.19 &	23.12 &	96.95 &	11.5 &	95.06 &	22.37 &	93.38 &	32.5 &	96.145 &	22.373 \\
1/4 &	99.20 &	1.93 &	96.77 &	13.37 &	94.83 &	24.37 &	 93.23 &	35.06 &	96.008 &	18.683 \\
1/2 &	99.43 &	1.87 &	96.97 &	14 &	95.15 &	22.75 &	93.07 &	35.68 &	96.155 &	18.575 \\
1/1 &	99.44 &	1.5 &	96.98 &	13.75 &	94.98 &	24.75 &	92.90 &	34.68 &	96.075 &	18.670 \\
2/1 &	99.46 &	1.5 &	97.06 &	14.62 &	95.06 &	22.87 &	92.91 &	32.93 &	96.123 &	17.980 \\
4/1 &	99.49 &	1.5 &	97.01 &	14 &	94.95 &	23.56 &	92.71 &	33.43 &	96.040 &	18.123 \\
8/1 &	99.53 &	1.5 &	97.01 &	13.43 &	94.85 &	22.75 &	92.57 &	33.18 &	95.990 &	17.715 \\
\hline
\end{tabular}
}}
    \end{table}%

\subsection{Performance of CLIPScope with Various Backbones}
\begin{figure*}[ht]
    \centering
    \includegraphics[width=0.24\linewidth]{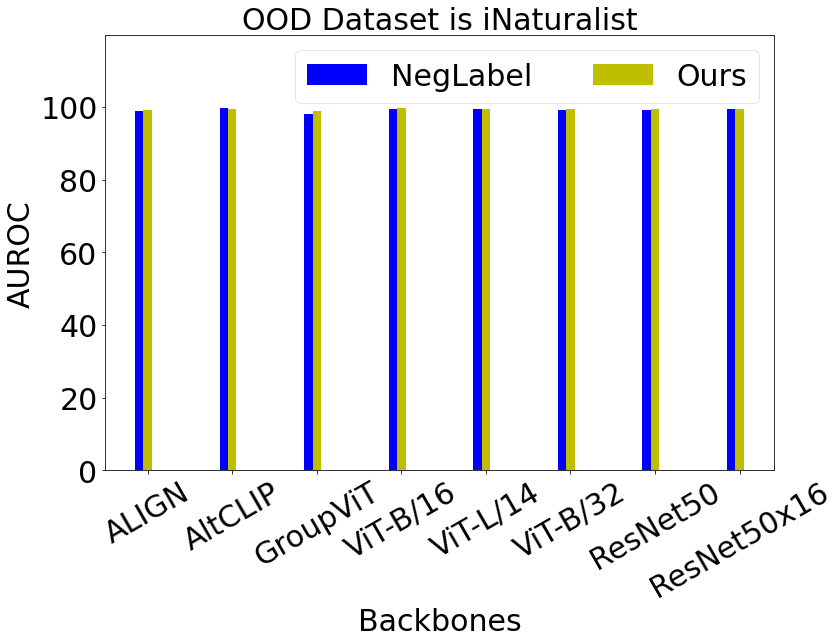}
    \includegraphics[width=0.24\linewidth]{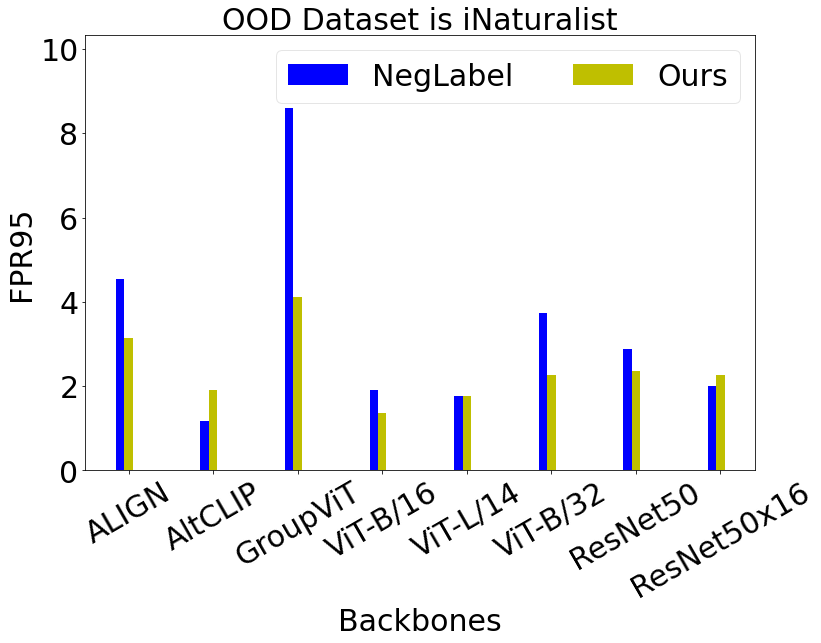}
    \includegraphics[width=0.24\linewidth]{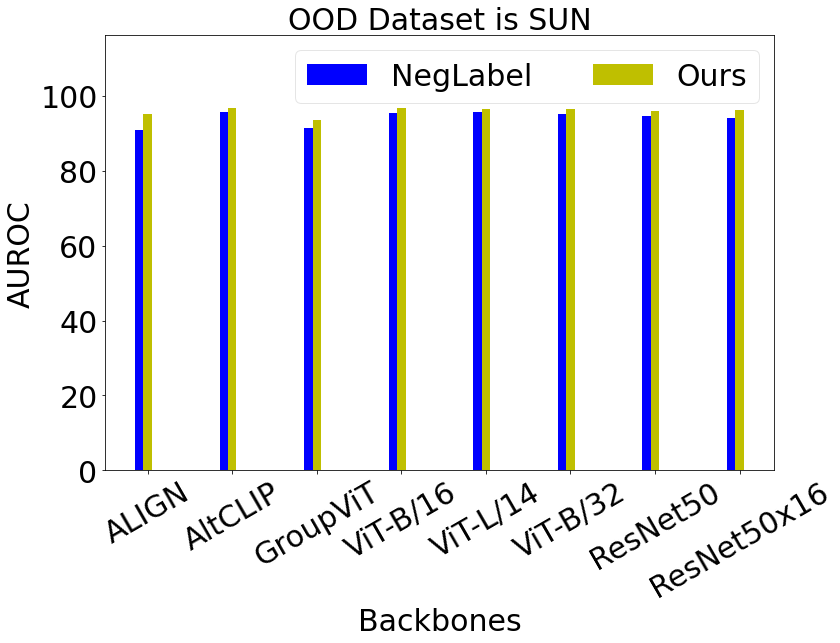}
    \includegraphics[width=0.24\linewidth]{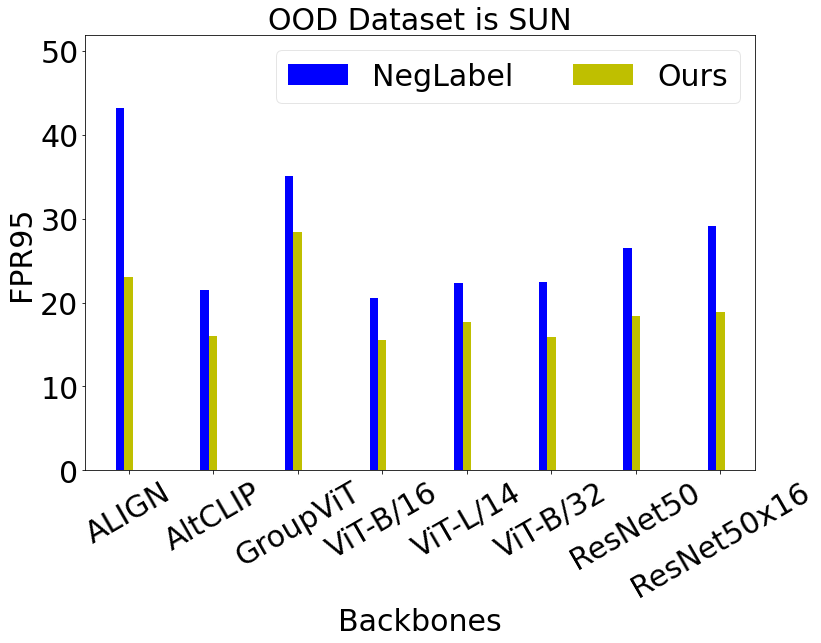}
    \includegraphics[width=0.24\linewidth]{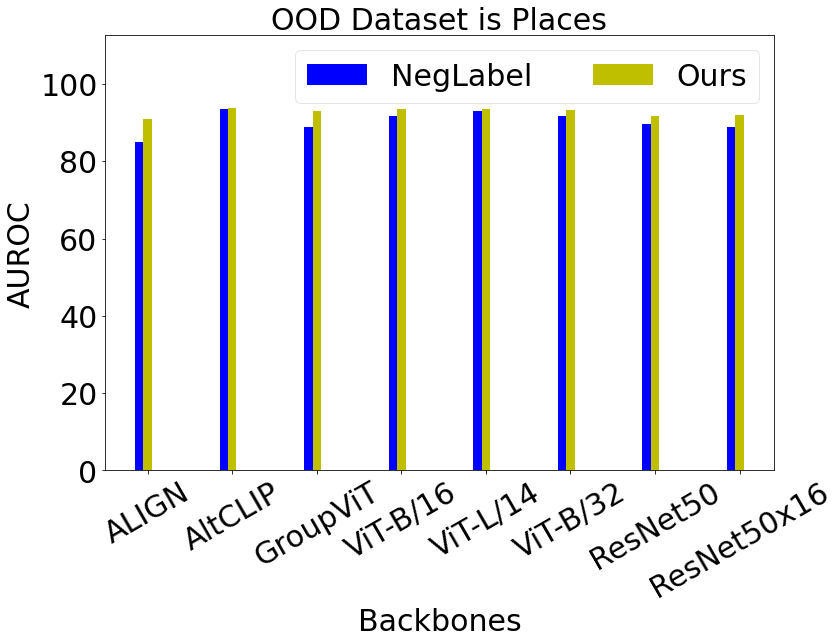}
    \includegraphics[width=0.24\linewidth]{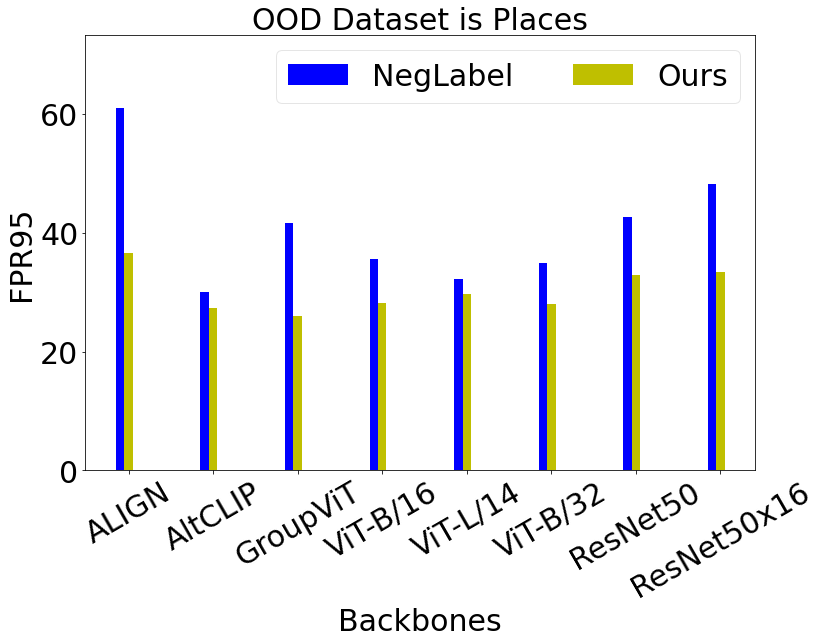}
    \includegraphics[width=0.24\linewidth]{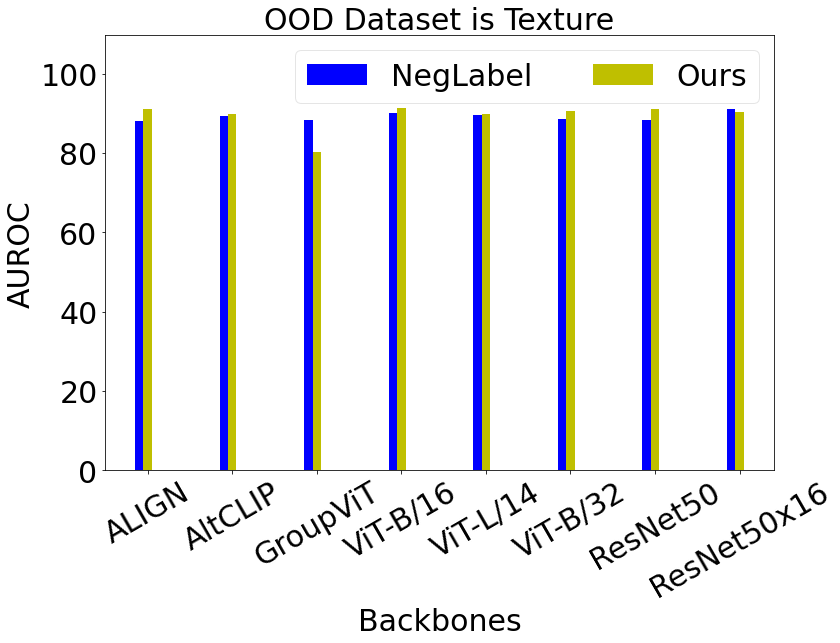}
    \includegraphics[width=0.24\linewidth]{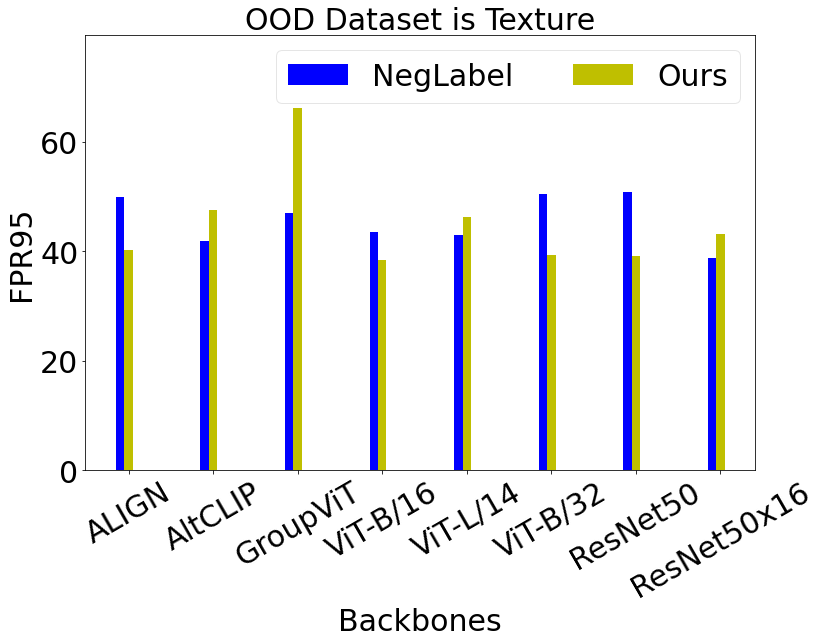}
    \caption{Performance (\%) of CLIPScope with various backbones. The ID dataset is ImageNet-1k.  }
    \label{fig:clip}
    %\vspace{-0.4cm}
\end{figure*}

We also evaluated the performance of CLIPScope using different backbones. Fig.\ref{fig:clip} shows the results.  Compared to NegLabel, CLIPScope consistently exhibits comparable or superior performance across most of the tested  models.  This diverse set of model evaluations demonstrates that CLIPScope is adaptable to different architectural frameworks. 

\subsection{Performance on Unbalanced Datasets}
\begin{table}
    \centering
    \textcolor{\colorname}{
        \caption{Performance of CLIPScope on unbalanced ID datasets. }
\label{tab:app_unbalance_class}
\resizebox{\textwidth}{!}{%
\begin{tabular}{@{}lcccccccccc@{}}
\toprule
OOD Dataset & \multicolumn{2}{c}{iNaturalist} & \multicolumn{2}{c}{SUN} & \multicolumn{2}{c}{Places} & \multicolumn{2}{c}{Textures} & \multicolumn{2}{c}{Average} \\ 
\cmidrule(r){2-11}
ID Datasets & AUROC$\uparrow$ & FPR95$\downarrow$ & AUROC$\uparrow$ & FPR95$\downarrow$ & AUROC$\uparrow$ & FPR95$\downarrow$ & AUROC$\uparrow$ & FPR95$\downarrow$ & AUROC$\uparrow$ & FPR95$\downarrow$ \\ 
%\addlinespace
ImageNet-1K & 99.60 &	1.28 &	97.34 &	13.52 &	94.20 &	26.32 &	93.04 &	34.41 &	96.045 &	18.883\\
\hdashline
ImageNet-10 &	98.98 &	5.11 &	97.41 &	9.66 &	92.46	 & 23.39 &	90.82 &	32.34 &	94.918 &	17.625 \\
ImageNet-20 &	99.61 &	1.50 &	96.43 &	15.01	 & 90.90	& 33.41 &	88.67 &	38.63 &	93.903 &	22.138 \\
ImageNet-100 &	99.35 &	2.14 &	95.75 &	17.15 &	90.70	& 31.34 &	87.46 &	44.02 &	93.315 &	23.663 \\
\hline
\end{tabular}
}}
\end{table}

\textcolor{\colorname}{
Table~\ref{tab:app_unbalance_class} presents the performance of CLIPScope on subsets of ImageNet, using ImageNet-1K as the ID labels. This setup creates an unbalanced class distribution, with some in-distribution classes having no samples. Our approach shows a slight decrease in AUROC and an increase in FPR95 in most cases due to this imbalance.  As discussed in the limitations section, potential misleading information, such as providing redundant ID labels, could negatively impact detection accuracy. However, as shown in Table~\ref{tab:app_smallID}, our approach is effective if the ID labels are correctly provided. }

\section{Further Discussions}
\subsection{Computation Complexity}

The computational complexity of CLIPScope is $\mathcal{O}(2MD)$ per image, where $M$ is the number of negative labels and $D$ is the dimension of the embedding feature. This complexity is the same as NegLabel's. Both methods use around 10,000 OOD labels and CLIP as the feature extractor, resulting in an efficient OOD detection time of about 1ms per sample. The mining algorithm, which processes large corpora like WordNet, takes only a few minutes on a single GPU machine and is performed before the inference phase, not affecting the inference speed. Importantly, CLIPScope calculates the confidence score for each input instance only once, eliminating the need for repeated scoring and improving computational efficiency. 

\subsection{Overlap Between Mined and Actual OOD Labels}

 Our OOD label mining strategy does not assume access to OOD test data, ensuring an unbiased selection of OOD labels without prior knowledge of the test data's OOD classes. This approach is similar to NegLabel's. Any overlap between the mined OOD labels and the actual OOD classes in the test data highlights the effectiveness of our mining strategy rather than being a drawback. NegLabel has previously justified using a wide range of concepts, potentially including the semantic labels of OOD samples, as a reasonable approach. This justification holds, especially when the corpus is large, similar to how vision-language models (VLMs) are considered suitable for evaluation in zero-shot tasks despite potential exposure to task-relevant data. When developers have specific insights into likely OOD labels, these can be intentionally included in the negative label space to further improve OOD detection effectiveness. Furthermore, Table~\ref{tab:mining} demonstrates that our approach remains effective even with a small number of $M$ (e.g., 0, 50, or 100). For smaller values of $M$, the mined OOD labels are less likely to overlap with actual OOD labels.

\subsection{ID Instances Classified into High Likelihood Classes}

\begin{figure}
    \centering
    \includegraphics[width=0.24\linewidth]{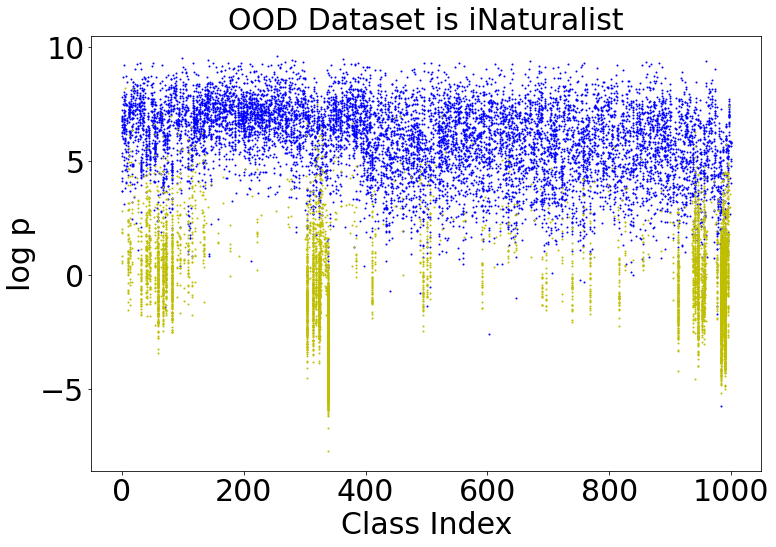}
    \includegraphics[width=0.24\linewidth]{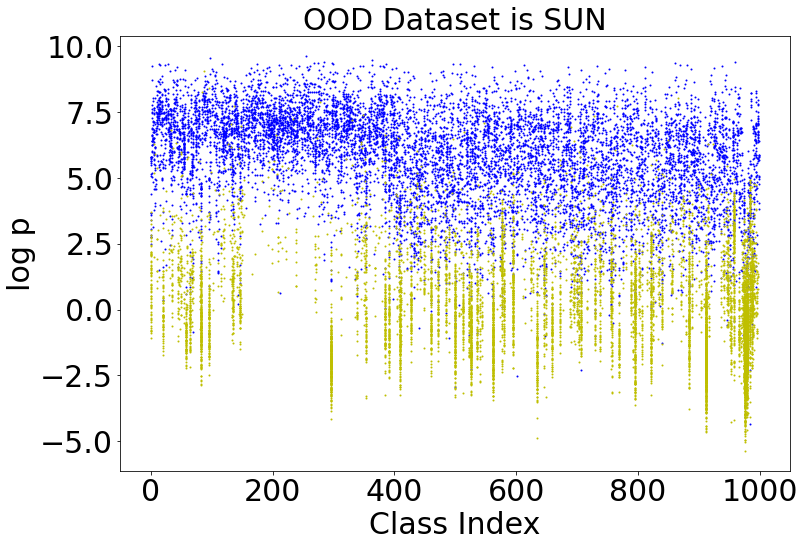}
    \includegraphics[width=0.24\linewidth]{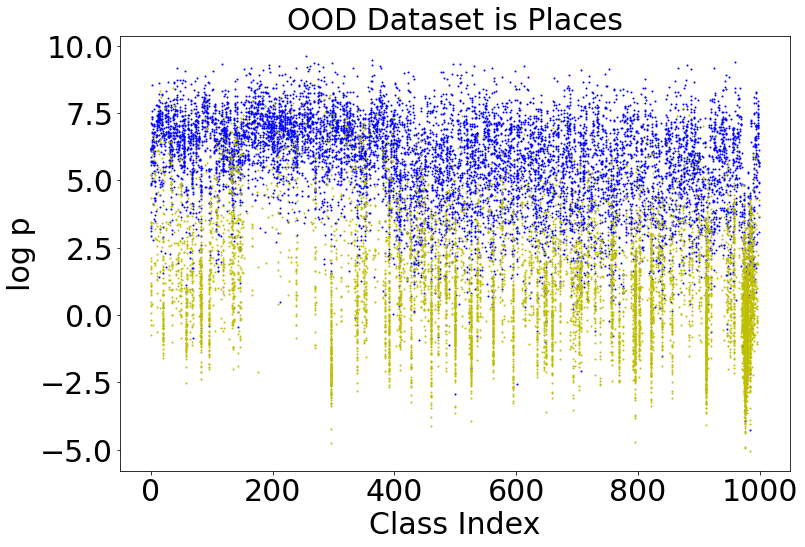}
    \includegraphics[width=0.24\linewidth]{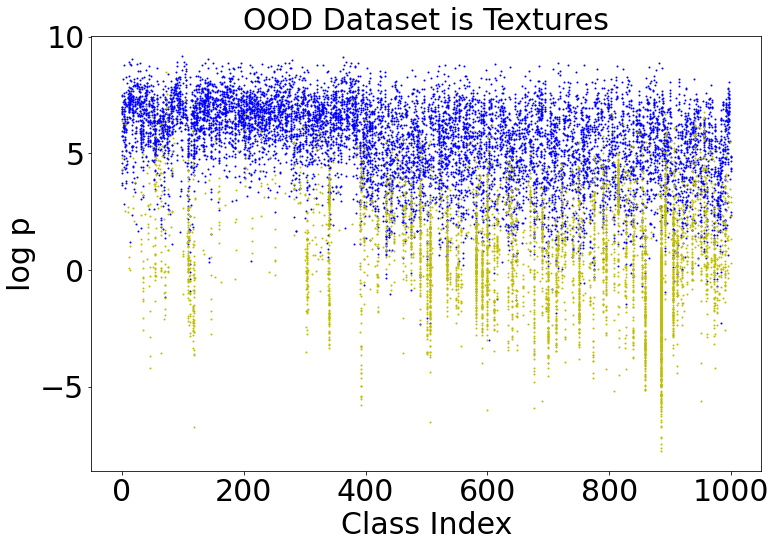}
    \caption{The logarithm confidence scores $\text{log } p$ of ID (blue) and OOD (yellow) samples.  The ID dataset is ImageNet-1k.}
    \label{fig:log}
\end{figure}

\begin{figure*}
\centering
    \includegraphics[width=0.19\linewidth]{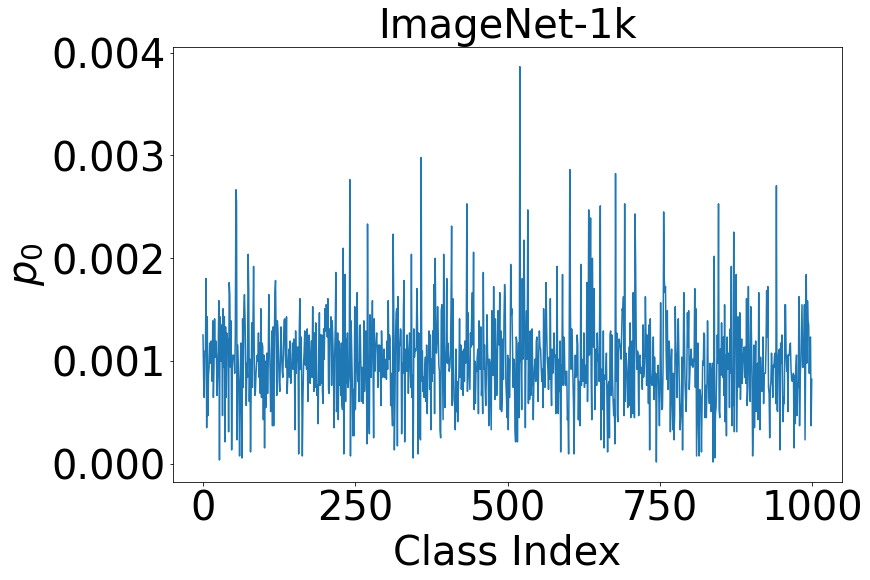}
    \includegraphics[width=0.19\linewidth]{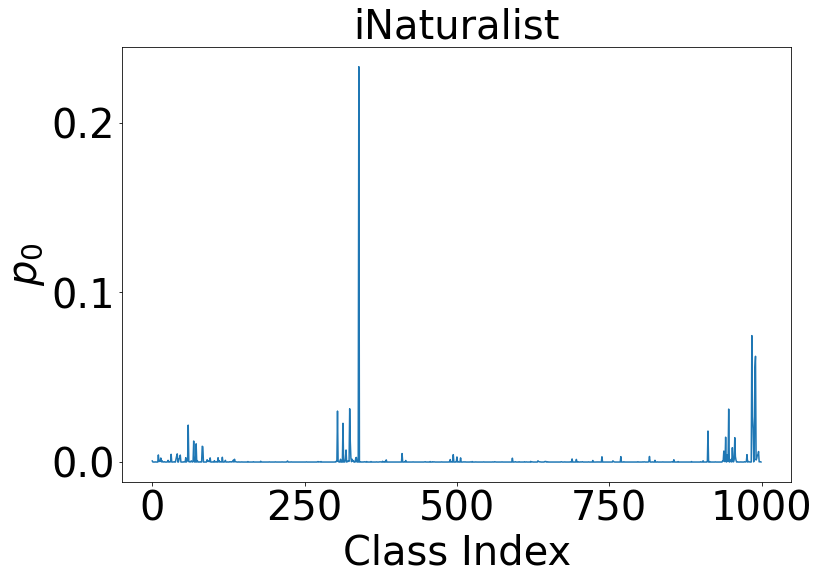}
    \includegraphics[width=0.19\linewidth]{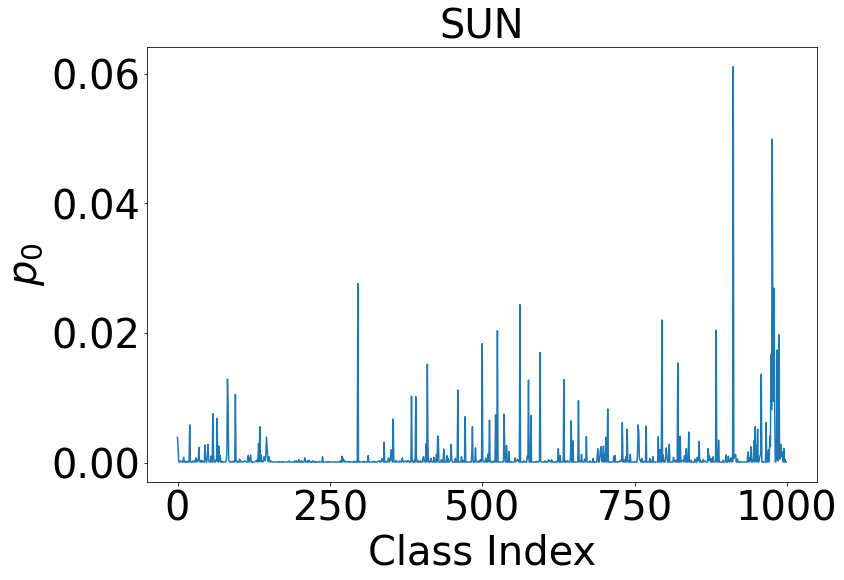}
    \includegraphics[width=0.19\linewidth]{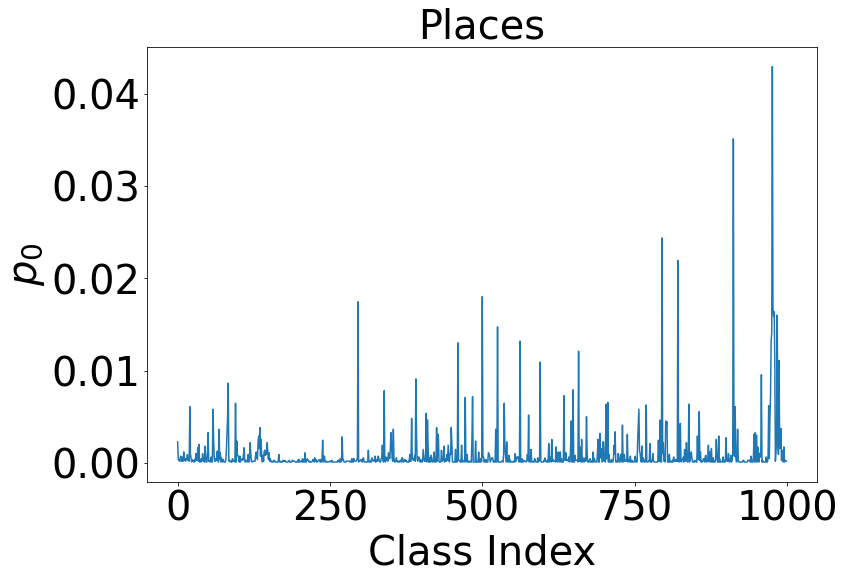}
    \includegraphics[width=0.19\linewidth]{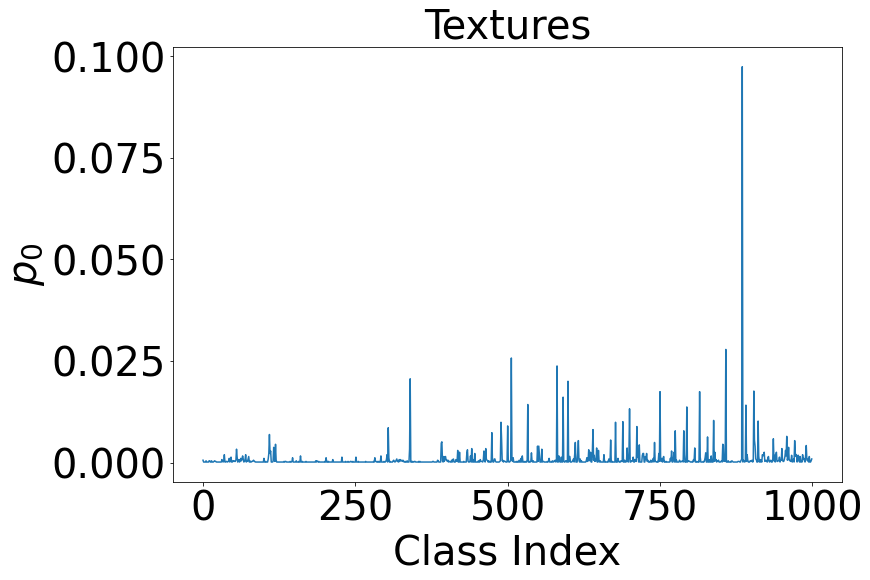}
    \caption{The classification behavior of CLIP on ID dataset is different from the classification behavior on OOD datasets. }
    \label{fig:likelihood}
\end{figure*}

ID instances that are classified into classes with high likelihood are influenced by the elevated class likelihood values. This effect is reflected in their confidence scores. Despite this influence, the confidence scores of these ID instances are still likely to surpass the threshold  because the numerator of their confidence scores is usually high. Indeed, Fig.~\ref{fig:log} shows the logarithm confidence scores $\text{log } p$ of ID and OOD samples for different datasets. Most ID instances have higher scores than OOD instances even in the high likelihood classes. Fig.~\ref{fig:likelihood} shows which classes have the highest likelihood.

The $p_0$ in Fig.~\ref{fig:likelihood} for each dataset $D$ is calculated as follows:
\begin{align}
    p_0(y_i) = \frac{1}{|D|} \sum_{x \in D} \mathds{1}(f(x, \mathcal{Y})=y_i) ~~~ \forall y_i \in \mathcal{Y}
\end{align} where $\mathcal{Y}$ is the ImageNet-1K labels. Given the ID dataset $D_I$ and the OOD dataset $D_O$, we have
\begin{align}
    \mathbb{P} (x\in\text{OOD} ~|~ f(x,\mathcal{Y})=y_i) = \frac{\sum_{x \in D_O} \mathds{1}(f(x, \mathcal{Y})=y_i)}{\sum_{x \in D_O} \mathds{1}(f(x, \mathcal{Y})=y_i)+\sum_{x \in D_I} \mathds{1}(f(x, \mathcal{Y})=y_i)}.
\end{align} Based on Fig.~\ref{fig:likelihood}, $ \mathbb{P} (x\in\text{OOD} ~|~ f(x,\mathcal{Y})=y_i)$ varies significantly between classes. However, CLIPScope provides very good performance in this general case, as evidenced by the results shown in Table~\ref{tab:comparison}.

\subsection{\textcolor{\colorname}{Training-Based Methods}}
\textcolor{\colorname}{
Training-based or tuning-based methods may improve their performance by using historical test samples. However, compared to training-based methods, our approach does not rely on ground-truth labels from historical test data and offers substantial advantages in terms of efficiency. It requires minimal memory, as it uses only histogram information based on the empirical output of CLIP rather than ground-truth labels, and it has faster computation compared to fine-tuning models with numerous parameters. These benefits make our method more practical for applications requiring frequent updates.}

\subsection{Broader Impact}
This paper presents work whose goal is to advance the field of machine learning. It demonstrates a  positive impact in the realm of zero-shot OOD detection by leveraging posterior information from historical instances. This approach has shown a considerable improvement in detection accuracy, setting a precedent for other OOD detection methods. The integration of posterior information into confidence score calculations could potentially enhance the performance of various OOD detection models, not limited to zero-shot approaches. However, the potential misleading information within the historical data  could adversely affect detection accuracy, compromising the reliability of open-world deployed machine learning systems.

\subsection{Future Works}
While CLIPScope currently utilizes only class likelihood as its form of posterior information, future explorations could delve into other types of posterior data. This expansion could uncover new dimensions of accuracy and efficiency in OOD detection. Furthermore, the development of new OOD detection scores remains a valuable and promising avenue of research. It would be  interesting to investigate how existing OOD detectors could benefit from the incorporation of CLIPScope's approach to using posterior information. 

\end{document}